\documentclass{article} 
\usepackage{iclr2025_conference,times}

\usepackage{amsmath,amsfonts,bm}

\def\eqref#1{equation~\ref{#1}}

\def\1{\bm{1}}

\DeclareMathAlphabet{\mathsfit}{\encodingdefault}{\sfdefault}{m}{sl}
\SetMathAlphabet{\mathsfit}{bold}{\encodingdefault}{\sfdefault}{bx}{n}

\usepackage{hyperref}
\usepackage{url}
\usepackage{booktabs}  
\usepackage{graphicx}
\usepackage{array}    
\usepackage{multirow}
\usepackage{multicol}
\usepackage{tablefootnote}
\usepackage{enumitem}
\usepackage{adjustbox}
\usepackage{mdframed}
\usepackage{subcaption}
\usepackage{xcolor}
\usepackage[ruled,vlined]{algorithm2e}
\mdfdefinestyle{MyFrame}{
    linecolor=black,
    linewidth=1pt,
    roundcorner=10pt,
    backgroundcolor=lightgray,
    innertopmargin=10pt,
    innerbottommargin=10pt,
    innerrightmargin=10pt,
    innerleftmargin=10pt,
    frametitlebackgroundcolor=gray!20,
    frametitlefont=\bfseries\itshape
}
\title{ReFeR: Improving Evaluation and Reasoning through Hierarchy of Models}

\author{
  \textbf{Yaswanth Narsupalli\thanks{Denotes that the first three authors contributed equally.}} \\
  IIT Kharagpur\footnotemark[2] \\
  \And
  \textbf{Abhranil Chandra\footnotemark[1]} \\
  University of Waterloo
  \And
  \textbf{Sreevatsa Muppirala\footnotemark[1]} \\
  IIT Kharagpur
  \AND
  \textbf{Manish Gupta} \\
  Microsoft, India
  \And
  \textbf{Pawan Goyal} \\
  IIT Kharagpur\thanks{
  Emails: \texttt{yasshu.yaswanth@gmail.com, pawang@cse.iitkgp.ac.in}
}
}

\newcommand{\mySpacing}{4pt}
\renewcommand\textfloatsep{\mySpacing}

\iclrfinalcopy 
\begin{document}

\maketitle

\begin{abstract}
Assessing the quality of outputs generated by generative models, such as large language models and vision language models, presents notable challenges. Traditional methods for evaluation typically rely on either human assessments, which are resource-intensive, or automatic metrics that often show a low correlation with human judgment. Another common approach is to use deep learning systems, which not only consume a substantial amount of compute and time but also require extensive training data. In this study, we introduce a tuning-free framework called ReFeR, designed to evaluate generative outputs, including both text and images, by leveraging a 2-level hierarchy of LLMs and VLMs themselves.  We rigorously evaluate our framework, ReFeR, across four diverse evaluation tasks. The framework not only improves the accuracy of these evaluations, surpassing previous benchmarks but also generates constructive feedback. Interestingly, the framework is also applicable to reasoning tasks. Experiments on four reasoning tasks demonstrate superior collective reasoning abilities of the framework. We present two variants of the framework: ReFeR-Turbo, optimized for accelerated performance, and ReFeR-Lite, offering a more cost-effective solution. ReFeR-Lite is $\sim7.7\times$ more efficient while being comparably accurate to ReFeR-Turbo. We make code, data and PIP package publicly available\footnote{{PIP}: \url{https://pypi.org/project/refer-agents/}}\footnote{
    {Git}: \url{https://github.com/yaswanth-iitkgp/ReFeR_Code}
}.

\end{abstract}

\section{Introduction}
\label{section:Intro}

The rapid production of content by large language models and vision language models (VLMs), poses a challenge to traditional human-centric evaluation methods and conventional automatic metrics. Metrics like BLEU \citep{BLUE}, ROUGE \citep{ROUGE}, and METEOR \citep{METEOR} for textual evaluation and CLIPScore \citep{clipscore} for image to text evaluation, often misalign with human judgment and face limitations in assessing creative or nuanced responses. Recent studies suggest using LLMs as novel, reference-independent evaluators by assessing text quality based on predicted sequence likelihoods, bypassing the need for direct reference comparisons \citep{chen-etal-2023-exploring-use}. This has motivated researchers~\citep{liu2023geval,chiang2023closer} to work on improving the evaluation capability of \emph{individual} LLMs on text evaluation.  \citet{zhang2024qualityassessmenteralarge} highlight that large models align more closely with human perceptual processes, thereby enhancing the evaluation of multimedia quality. Consequently, \citet{chen2023xiqeexplainableimagequality} leverage vision language models to provide explainable image quality evaluation by generating textual explanations, assessing fidelity, alignment, and aesthetics.

Surprisingly, despite the potential for improved performance by using ensembles of multiple vision-language models or large language models, there has been limited research on how to align evaluations from multiple VLMs or LLMs with human judgments. While the concept of using multiple VLMs or LLMs together to solve this complex problem is promising, it introduces several uncertainties, including how to select the models, how many models to use, how to manage communication between different models and what prompting structure should be used to maximize the effect. 

In this paper, we introduce a multi-agent Reason-Feedback-Review (ReFeR) framework, drawing inspiration from the academic peer review process to enhance the evaluation of multimodal generative outputs like text generated by an LLM, an image generated by any model,

or caption of an image generated by a VLM. By using multiple LLMs or VLMs as evaluators and feedback providers in a system akin to academic peer review, ReFeR enables a comprehensive evaluation of generative outputs across various domains, promoting model self-improvement, explainability, and robustness in complex scenarios. The paper outlines ReFeR's methodology, including its new prompting schema and the strategic use of LLMs or VLMs in roles parallel to peer reviewers and area chairs, facilitating a multi-dimensional evaluation through a hierarchical framework consisting of two levels: evaluation at the peer level and evaluation at the area chair level.

The framework is tested across two NLG evaluation and two multimodal evaluation tasks. Interestingly, the framework is generic enough to be applicable for other tasks beyond evaluation. Hence, we also test the framework's reasoning ability on four reasoning benchmarks. Furthermore, ReFeR's feedback mechanism has enabled the production of instruction-tuning datasets, which can be used to fine-tune smaller models achieving a better correlation with human evaluation.

We present two variants of our proposed framework, ReFeR Turbo and ReFeR Lite. ReFeR Lite is $\sim$7.7$\times$ more efficient than ReFeR Turbo. Both the variants outperform strong baselines on both text evaluation datasets: TopicalChat~\citep{mehri2020usrunsupervisedreferencefree} and SummEval~\citep{fabbri2021summeval}. ReFeR also beats baselines like Clipscore~\citep{clipscore}, ImageReward~\citep{xu2023imagerewardlearningevaluatinghuman} and others on caption quality and image generation quality evaluation using ICQD~\citep{icqd2019} and AGIQA~\citep{zhang2023perceptualqualityassessmentexploration} datasets respectively by large margins. Lastly, ReFeR also beats single agent methods (zero-shot CoT~\citep{kojima2023largelanguagemodelszeroshot}, self correction~\citep{huang2024largelanguagemodelsselfcorrect}), and multi-agent methods like multi-agent debate~\citep{du2023improvingfactualityreasoninglanguage} and multi-agent peer review~\citep{xu2023reasoninglargelanguagemodels} on 3 out of 4 reasoning datasets, clearly outperforming on average while keeping lower costs than baselines.  

To summarize, the primary contributions of our research are as follows:
    (1) Introduction of a general-purpose hierarchical framework, called ReFeR, given in two variants, ReFeR-Turbo and ReFeR-Lite.
    (2) We develop a novel prompting schema, with a novel eval guidelines component, specifically designed to improve the effectiveness of our framework in evaluation and reasoning tasks. 
    (3) Empirical validation of the framework's evaluation and reasoning skills on four benchmarks each. 
    (4) We conduct an in-depth analysis of our multi-agent framework, ReFeR, addressing key questions such as how to select models, how many models to use, and other critical aspects of model interaction.
    
\section{Methodology}
\label{section:Methodology}

In this section, we present the ReFeR framework, its mathematical formulation, the prompting schema, and its variants.

\subsection{ReFeR Framework}
\label{refer-framework}

\begin{figure*}[!t]
\centering\includegraphics[width=\textwidth]{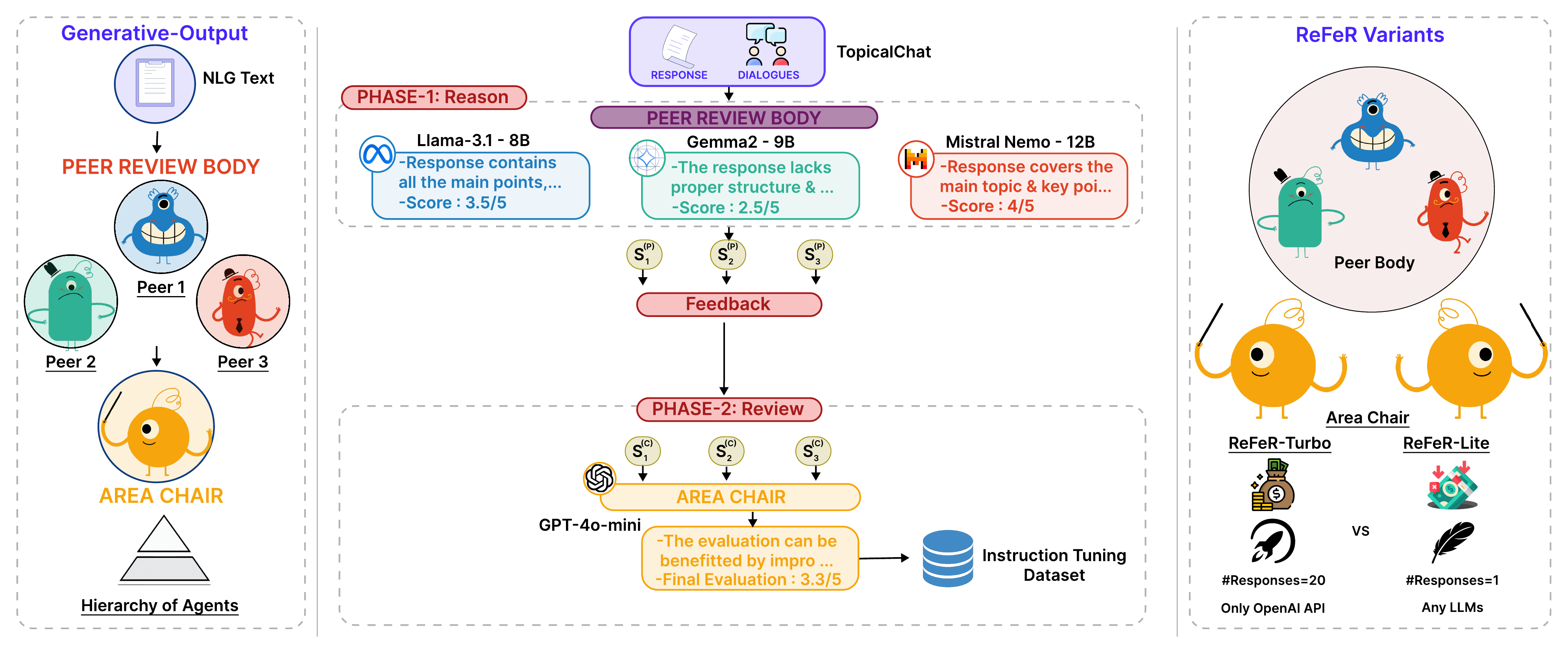}
    \caption{\textbf{Illustration of the ReFeR Framework on the TopicalChat dataset.} Refer to Fig. \ref{fig:refer-multimodal} (in the appendix) for illustration of ReFeR for multimodality and Algorithm~\ref{algo:refer} showing the framework's working. We use the predictions from AC to create an Instruction tuning dataset which can be used to improve the performance of smaller models as evaluators, shown in Appendix~\ref{appendix: finetuning}} 
    \label{fig:ReFeR}
\end{figure*}

Evaluating generative outputs without a predefined correct answer, such as assessing the quality of a research paper or open-ended responses, presents significant challenges. Inspired by the hierarchical peer review process in academia, we propose the ReFeR framework, which leverages a hierarchy of language models to systematically evaluate generative outputs.
The ReFeR framework consists of two main modules, as depicted in Figure~\ref{fig:ReFeR}. 

\textbf{1. Peer Review Body}

Let $G$ denote the generative output to be evaluated, $E_P$ represent the prompt for the peer and $E_{AC}$ represent the prompt for the area chair. 

Let $\mathcal{P} = \{P_1, P_2, \dots, P_K\}$ be a set of $K$ peer agents, where each $P_i$ is a language model acting as a peer reviewer. Each peer agent independently evaluates $G$ according to $E_P$, producing a comment $C_i$ and a score $S_i \in \mathbb{R}$. This process is formalized as follows.
\begin{equation}
(C_i, S_i) = \text{Evaluate}_{P_i}(G, E_P), \quad \forall i \in \{1, 2, \dots, K\}
\end{equation}
\textbf{2. Area chair Evaluation}

An area chair agent $AC$, typically a larger or more capable language model, synthesizes the peer reviews to provide the final evaluation. The area chair considers the generative output $G$, the prompt $E_{AC}$, and the set of peer reviews $\{(C_i, S_i)\}_{i=1}^K$, producing a final comment $C_{\text{final}}$ and a final score $S_{\text{final}}$. \textit{n} is a hyperparameter that denotes the number of responses for a given prompt.
\begin{eqnarray}
\{(C_{\text{AC}}^{(j)}, S_{\text{AC}}^{(j)})\}_{j=1}^n &=& \left\{ \text{Evaluate}_{AC}^{(j)}(G, E_{AC}, \{(C_i, S_i)\}_{i=1}^K) \right\}\\
\label{eq:eq2}
S_{\text{final}} &=& \frac{1}{n} \sum_{j=1}^n S_{\text{AC}}^{(j)}
\end{eqnarray}

\subsection{Prompting Schema}
\label{prompting-schema}

An essential aspect of assessing generative outputs with language model agents involves crafting prompts that elicit high-quality evaluations. Prior work, such as G-Eval by \citet{liu2023geval}, introduced a structured evaluation schema, organizing the prompt into sections: task introduction, evaluation criteria, steps for evaluation, input presentation (context and target), and an evaluation form designed to output a numerical rating only. Subsequently, \citet{chiang2023closer} proposed an adjusted schema named Analyze-Rate, which prioritizes an analytical review followed by scoring, showing improved performance over G-Eval's prompt. 

To further refine this approach, we introduce a new module in the evaluation schema called \emph{Evaluation Guidelines} to enhance the model's understanding of the scoring criteria, akin to guidelines provided in traditional academic review processes. Evaluation guidelines can be automatically generated by prompting a language model with the prompt structure and some examples from the dataset and we call this process auto prompt. We give an example of this process in Appendix~\ref{appendix:Auto-prompt} showing the \textit{Auto Prompt} for Engagingness prompt for TopicalChat. 

Alternatively, manually written human annotation guidelines of the dataset can be used. We also modified the evaluation form to include a critical comment or reasoning for the given score. The proposed evaluation schema is shown in Figure~\ref{fig:eval-schema}.

\subsection{ReFeR Variants}
\label{sec:refer-variants}

To enhance the framework's versatility and performance, we developed two variants of ReFeR: \textbf{ReFeR-Turbo} and \textbf{ReFeR-Lite}.

\subsubsection{ReFeR-Turbo}

ReFeR-Turbo leverages a hyperparameter $n$, representing the number of responses generated by the area chair agent. This variant generates multiple responses ($n = 20$) for each prompt, applying a scoring function that averages the scores across all generated responses, as described in Eq.~\ref{eq:eq2}.

The final comment $C_{\text{final}}$ is the list of all individual comments from the area chair evaluations. While ReFeR-Turbo provides superior performance due to generating more evaluations per prompt, it incurs higher computational costs. Additionally, the use of $n=20$ is often constrained to models from the OpenAI API, as other APIs supporting large models do not support this level of multiple response generation directly. Although it is possible to generate multiple responses by making repeated calls to the model (e.g., running the model 20 times with the same prompt), this approach is computationally expensive and less practical for large-scale evaluation tasks. This usage of the hyperparameter was first suggested by G-Eval and later used by Analyze-Rate.

\subsubsection{ReFeR-Lite}

To enhance flexibility and reduce computational overhead, we developed ReFeR-Lite, which removes the dependency on the parameter $n$ for the given performance. In this variant, only a single response ($n = 1$) is generated for each prompt, or $n$ is completely removed. This reduction in response generation is reflected in Eq 2, where $n$ is set to 1.
\begin{equation}
(C_{\text{final}}, S_{\text{final}}) = \text{Evaluate}_{AC}(G, E, \{(C_i, S_i)\}_{i=1}^K).
\end{equation}
By relying on just one evaluation per prompt, ReFeR-Lite can be used with a wider variety of models, including open-source models, which do not natively support the generation of multiple responses with a single prompt. Despite generating fewer responses, ReFeR-Lite maintains competitive performance and offers significant cost savings. This makes it an efficient and cost-effective solution for tasks where computational resources are limited or where evaluating large numbers of samples is required.

Both ReFeR-Turbo and ReFeR-Lite use the same peer evaluation structure, but differ primarily in the area chair's response generation and model compatibility. ReFeR-Turbo, with $n=20$, offers potentially higher performance due to generating more evaluations but is restricted to models that support or can simulate multiple response generation with a single prompt. In contrast, ReFeR-Lite provides greater flexibility and cost-efficiency by generating only a single response ($n=1$) per prompt, making it more suitable for resource-constrained environments.

\section{Experiments}
\label{section:Experiments}

\subsection{Datasets}
\label{subsection:datasets}

For NLG evaluation, we test our framework on SummEval \citep{fabbri2021summeval} for summarization evaluation, and TopicalChat \citep{mehri2020usrunsupervisedreferencefree} for dialogue generation evaluation. For multimodal evaluation, we compare our framework on evaluating two types of task, image-to-text using ICQD (Image Caption Quality Dataset)~\citep{icqd2019} and text-to-image generation using AGIQA-1k by \citet{zhang2023perceptualqualityassessmentexploration}. For ICQD, we score model-generated captions and compare them with the average human annotated rating for the same. In AGIQA, we assess the quality of AI-generated images in reference to a given prompt and compare it with the mean opinion score (human annotations).

We also test our framework on 4 reasoning datasets: AQuA \citep{ling-etal-2017-program}, BBH-DU \citep{srivastava2023imitationgamequantifyingextrapolating}, CSQA \citep{aggarwal-etal-2021-explanations} and GSM8k \citep{cobbe2021trainingverifierssolvemath} which cover various reasoning tasks like Math, Commonsense and Date Understanding. Statistics and details about all the datasets are provided in Table \ref{tab:Datasets}. For more details about the datasets, refer to Appendix \ref{appendix:datasets}. We test our framework on these reasoning tasks, where our framework answers a reasoning question with the label or numerical Answer after giving the reasoning. We calculate the accuracy of our answers in reference to the gold answers.

\begin{table}[ht]
\caption{\textbf{Dataset Statistics}. We list all the tasks we tackle in our paper and the datasets we used to show results with the number of samples used.\tablefootnote{For Reasoning, a random subset of 100 was sampled from the original datasets, following \citep{chen2024reconcile}. 500 random samples were selected from the original AGIQA-1k to get a well-distributed dataset. We use 864 samples with usable image urls from the ICQD test dataset. We use the full test sets for the NLG Evaluation datasets.}}
\centering
\scriptsize
\begin{tabular}{|l| l| l| c| l| l|}
\hline
\textbf{Dataset} & \textbf{Domain} & \textbf{Task} & \textbf{Samples} & \textbf{Answer} & \textbf{Scale} \\
\hline
TopicalChat & Dialogue Generation & NLG Evaluation & 360 & Rating (on 4 metrics) & 1-3 \\
SummEval & Summarization & NLG Evaluation & 1600 & Rating (on 4 metrics) & 1-5 \\
ICQD & Image-to-Text & Multimodal Evaluation & 864 & Caption Score & 0-100\\
AGIQA & Text-to-Image & Multimodal Evaluation & 500 & Generation Score & 0-5\\
AQuA & Math & Reasoning & 100 & Option & A-E\\
CSQA & Commonsense & Reasoning & 100 & Option & A-E\\
BBH-DU & Date Understanding & Reasoning & 100 & Option & A-F\\
GSM8k & Math & Reasoning & 100 & Number & \textbf{-}\\
\hline
\end{tabular}
\label{tab:Datasets}
\end{table}

\subsection{Baselines} 
\label{subsection:baselines}

\subsubsection*{NLG Evaluation}

While the current landscape of models for evaluating NLG responses includes reference-based methods such as BERTScore \citep{zhang2020bertscore}, UniEval \citep{zhong2022unifiedmultidimensionalevaluatortext} and reference-free methods like GPTScore \citep{fu2023gptscore}, we do not consider these models as baselines given they were clearly surpassed by G-Eval
 \citep{liu2023geval} and later works \citep{chiang2023closer}. Given our work primarily proposes a reference-free LLM-based evaluation for NLG, we do a comparative analysis primarily against G-Eval \citep{liu2023geval} and Analyze-Rate \citep{chiang2023closer} only.

\textbf{G-Eval} performs evaluation by deploying a single LLM agent. This agent employs Auto-CoT (chain of thought) reasoning and a form-filling paradigm to ascertain the quality of NLG outputs, delivering only scores for the specific dimensions under scrutiny.  \\
\textbf{Analyze-Rate} builds upon G-Eval, advocating for an enhanced prompt
structure. This methodology incorporates a preliminary analysis phase before scoring, aiming to enrich the evaluative process for NLG tasks.

\subsubsection*{MultiModal Evaluation}

For multimodal evaluation, several works like HyperIQA \citep{Su_2020_CVPR}, DBCNN \citep{Zhang_2020}, IP-IQA \citep{qu2024bringingtextualpromptaigenerated} were proposed for image quality assessment, but all of these works are deep learning-based methods which leverage and depend on training a capable model. Hence we do not compare our framework against them directly. 

\textbf{CLIP Score} \citep{clipscore} evaluates how well an image aligns with a text description by using the CLIP model, which computes similarity scores between images and text embeddings.\\
\textbf{Image Reward} \citep{xu2023imagerewardlearningevaluatinghuman} is a scoring model trained to assess the quality/alignment of generated images with text by comparing them against reference images using a reward model.\\
\textbf{Pick Score} \citep{kirstain2023pickapicopendatasetuser} is another scoring model for the task of image text alignment, which is trained on human preference images `picked' for a given text.\\
\textbf{X-IQE} \citep{chen2023xiqeexplainableimagequality} leverages VLMs to evaluate text-to-image generation methods by generating textual explanations. We implement their Alignment dimension experiments to compare with our results on text-to-image generation dataset (AGIQA).

\subsubsection*{Reasoning}

We compare our framework against a variety of baseline methods across different categories. For single-agent methods, we select zero-shot Chain-of-Thought (CoT) and Self-Correct. For multi-agent frameworks, we compare against Multi-Agent Debate and Multi-Agent Peer Review, both of which use a single model acting as multiple agents. 

\textbf{Zero-shot CoT} \citep{kojima2023largelanguagemodelszeroshot} utilizes chain-of-thought prompting to generate reasoning processes and answers using a single agent.\\
\textbf{Self-Correct} \citep{huang2024largelanguagemodelsselfcorrect} is a single-agent approach that enables an LLM to iteratively evaluate its own outputs, identify errors, and refine its responses through self-reflection.\\
\textbf{Multi-Agent Debate} \citep{du2023improvingfactualityreasoninglanguage} involves a group of agents, where each agent observes the solutions provided by others, updates its own solution accordingly, and repeats this process through multiple iterations.\\
\textbf{Multi-Agent Peer Review} \citep{xu2023reasoninglargelanguagemodels} is a multi-agent system in which each agent independently generates a solution, reviews the solutions of others, and assigns confidence scores to its reviews. Agents then revise their initial solutions based on the received peer reviews. This revision is repeated through multiple iterations/rounds of peer review. We used the default number of rounds (3) mentioned by the authors.

\subsection{Implementation Details}
\label{subsection:implement-detail}

\textbf{NLG Evaluation:} Our framework for NLG evaluation employs Llama-3.1-8B-Instruct \citep{llama31_8b_instruct}, Mistral-Nemo-12B \citep{mistral_nemo_instruct} and Gemma-2-9B \citep{gemma2_9b} as the peer models and GPT-4o-mini \citep{gpt4o_mini} as the area chair model. We use \citet{together-ai}'s API for the peer models, but since these are small open-source models, they can also be deployed locally. For the baselines, we follow the original setups proposed by \citet{liu2023geval} and \citet{chiang2023closer}. As mentioned in Section \ref{sec:refer-variants}, we vary the hyperparameter \textit{n} for the two ReFeR variants. For more details on other hyperparameters, refer to Appendix \ref{appendix:hyper-param}.

\textbf{Multimodal Evaluation:} For multimodal evaluation, our framework uses only 2 peers: Gemini-1.5-Flash \citep{gemini_1.5_flash} and GPT-4o-mini \citep{gpt4o_mini}. We use GPT-4o \citep{gpt4o} as the area chair model. We choose only 2 peers for multimodal evaluation setup considering the cost and availability of VLMs of similar strength. More details on the number of peers and how to choose peers are described in Section \ref{section: analysis}. The baselines like CLIPScore \citep{clipscore}, ImageReward \citep{xu2023imagerewardlearningevaluatinghuman}, PickScore \citep{kirstain2023pickapicopendatasetuser} are implemented following the codes provided in their official repositories. 

\textbf{Reasoning:} We use the same setup as our NLG evaluation for all our reasoning experiments following similar prompting structure except using evaluation guidelines which is irrelevant in reasoning tasks. All the baselines were implemented and evaluated using the scripts provided by \citet{xu2023reasoninglargelanguagemodels} in their official repository.

\section{Results and Discussions}
\label{section:results-discussion}
This section presents the experimental results evaluating ReFeR's effectiveness in assessing text, multimodal outputs, and reasoning capabilities. Experimental details are provided in Section \ref{subsection:implement-detail}, hyperparameters in Appendix \ref{appendix:hyper-param}, and prompts in Appendix \ref{appendix:Prompts}.

\subsection{NLG Evaluation}
\label{subsection:nlg-eval-results}
We evaluate ReFeR's performance on two datasets: TopicalChat and SummEval. For TopicalChat, we assess dialog system responses based on four metrics: Coherence, Engagingness, Groundedness, and Naturalness. For SummEval, we evaluate article summaries using Coherence, Consistency, Fluency, and Relevance metrics. Following \citep{liu2023geval} and \citep{chiang2023closer}, we compare the generated scores with human-annotated ground truth using Spearman $(\rho)$ and Kendall-tau $(\tau)$ correlations.
Results for TopicalChat are presented in Table \ref{tab:NLG-TopicalChat}, with SummEval results in Appendix \ref{appendix:SummEval}. All results are averaged over three runs. The table first shows individual peer performances using our peer prompt, followed by baselines Analyze-Rate \citep{chiang2023closer} and G-Eval \citep{liu2023geval}.
ReFeR Turbo outperforms all baselines on most metrics and excels on average. ReFeR Lite, our cost-effective model, ranks second on average despite generating a single response instead of 20 like G-Eval and Analyze-Rate. G-Eval sometimes outperforms Analyze-Rate despite only generating scores, while both Analyze-Rate and ReFeR provide analysis in addition to scores, offering the potential for model improvement. The key findings from this experiment are:

(1) Both ReFeR Turbo and ReFeR Lite outperform baselines.
(2) ReFeR Lite with $n=1$ also achieves comparable performance which being significantly cheaper.

While generating multiple responses (e.g., $n=20$ as in G-Eval) is theoretically possible with any LLM, it poses substantial practical challenges. For instance, evaluating the TopicalChat dataset (360 samples, 4 metrics) would require approximately 28,800 model calls with an average input token size of 675 for TopicalChat. This approach becomes impractical in terms of cost, time, and computational resources, especially for models without the throughput ($n=20$) capabilities of the OpenAI API. Hence ReFeR-Lite can be an option in such cases.

\begin{table}[ht!]
\centering
\scriptsize
\caption{\textbf{Comparison of ReFeR with baselines for NLG evaluation on the TopicalChat dataset.} Results are averaged across 3 runs. The best results are bolded, and the second-best are underlined. *Costs for ReFeR Turbo and ReFeR Lite include only AC API cost, as open-source peer models can be deployed locally and so do not involve API costs. Peer model costs based on API pricing from services like \citep{together-ai} are also provided for reference. Relative costs are shown as fractions of the most expensive method.}
\label{tab:NLG-TopicalChat}
\begin{tabular}{|c|c|c|c|c|c|c|c|c|c|c|c|c|}
\hline
\textbf{} & \multirow{2}{*}{\textbf{Method}} & \multicolumn{2}{c|}{\textbf{Coherence}} & \multicolumn{2}{c|}{\textbf{Engagingness}} & \multicolumn{2}{c|}{\textbf{Groundedness}} & \multicolumn{2}{c|}{\textbf{Naturalness}} & \multicolumn{2}{c|}{\textbf{Average}} & \textbf{Cost} \\
\cline{3-13}
& & $\rho$ & $\tau$ & $\rho$ & $\tau$ & $\rho$ & $\tau$ & $\rho$ & $\tau$ & $\rho$ & $\tau$ & (Relative) \\
\hline
\multirow{3}{*}{Peer Agents} & Llama-3.1-8B & 0.380 & 0.324 & 0.400 & 0.342 & 0.444 & 0.414 & 0.320 & 0.268 & 0.386 & 0.337 & 0.13 \\
& Mistral Nemo-12B & 0.409 & 0.346 & 0.594 & 0.501 & 0.442 & 0.414 & 0.411 & 0.348 & 0.464 & 0.402 & 0.23 \\
& Gemma-2-9B & 0.536 & 0.453 & 0.615 & 0.527 & 0.582 & 0.545 & 0.519 & 0.430 & 0.563 & 0.489 & 0.20 \\
\hline
\multirow{2}{*}{Baselines} & Analyze-Rate & 0.505 & 0.393 & \underline{0.647} & 0.503 & 0.463 & 0.388 & 0.572 & 0.442 & 0.547 & 0.432 & 0.77 \\
& G-Eval & \textbf{0.587} & \textbf{0.474} & 0.444 & 0.384 & 0.526 & 0.506 & \underline{0.599} & 0.468 & 0.539 & 0.458 & 0.13 \\
\hline
\multirow{2}{*}{Ours} & ReFeR Turbo & \underline{0.585} & \underline{0.454} & \textbf{0.673} & \textbf{0.535} & \textbf{0.628} & \textbf{0.577} & \textbf{0.625} & \textbf{0.484} & \textbf{0.628} & \textbf{0.513} & 1.0$^{*}$ \\
& ReFeR Lite & 0.535 & 0.452 & 0.624 & \underline{0.533} & \underline{0.583} & \underline{0.546} & 0.575 & \underline{0.482} & \underline{0.579} & \underline{0.503} & 0.13$^{*}$ \\
\hline
\end{tabular}
\end{table}

\subsection{Multimodal Evaluation}
\label{subsection:multi-modal}
To assess the multimodal applicability of ReFeR, we conducted experiments on two tasks: image generation quality evaluation using the AGIQA dataset (text-to-image setting) and image caption evaluation using the ICQD dataset (image-to-text setting). Table \ref{tab:image-quality-combined} presents the results of these experiments.
Following previous deep learning-based works such as \citep{zhang2023perceptualqualityassessmentexploration}, we report Spearman's $\rho$ and Kendall's $\tau$ rank correlations. Key findings include the following.
\begin{itemize}[noitemsep, left=0em]
\item ICQD dataset: Both variants of ReFeR outperform all baselines. Notably, although individual peers show low correlations, AC effectively countered this, resulting in better correlation.
\item AGIQA dataset: ReFeR Turbo outperforms all baselines, while ReFeR Lite outperforms ClipScore and X-IQE but falls short of ImageReward and PickScore.
\end{itemize}
We attribute the performance difference in the AGIQA dataset to the fact that both ImageReward and PickScore involve training based on human preferences, which may have contributed to their superior performance compared to our ReFeR Lite variant. But, our ReFeR-Lite has clearly surpassed a single VLM based method X-IQE by a large margin showing the effectiveness of the framework.

\begin{table}[ht!]
\centering
\scriptsize
\caption{\textbf{Multimodal Evaluation Results.} Comparison of caption quality and image generation quality score correlations with human scores on ICQD and AGIQA datasets, respectively. *X-IQE is a text-to-image VLM-based method, so we don't show it for Caption Quality. 
}
\label{tab:image-quality-combined}
\begin{tabular}{|l|l|c|c|c|c|c|c|c|}
\hline
& \multirow{2}{*}{\textbf{Method}} & \multicolumn{2}{c|}{\textbf{Caption Quality}} & \multicolumn{2}{c|}{\textbf{Image Quality}} &\textbf{Cost} \\
\cline{3-7}
 &  & \textbf{$\rho$} & \textbf{$\tau$} & \textbf{$\rho$} & \textbf{$\tau$} & (Relative) \\
\hline
\multirow{2}{*}{Peer Agents} 
    & Gemini-1.5-Flash & 0.135 & 0.098 & 0.341 & 0.268 & 0.07\\
    & GPT-4o-mini & 0.200 & 0.145 & 0.502 & 0.392 & 0.01\\
\hline
\multirow{3}{*}{Baselines} 
    & Clip Score & 0.310 & 0.233 & 0.522 & 0.366 & -\\
    & ImageReward & 0.433 & 0.302 & \underline{0.634} & \underline{0.451} & -\\
    & Pick Score & 0.352 & 0.241 & 0.627 & 0.442 & -\\
    & X-IQE* & - & - & 0.410 & 0.307 & 0.05\\
\hline
\multirow{2}{*}{Ours} 
    & ReFeR Turbo & \textbf{0.497} & \textbf{0.347} & \textbf{0.657} & \textbf{0.467} & 1.0\\
    & ReFeR Lite & \underline{0.459} & \underline{0.336} & 0.599 & 0.442 & 0.14\\
\hline
\end{tabular}
\end{table}

\subsection{Reasoning}
\label{subsection: reasoning}
We hypothesize that our framework enhances the overall reasoning capabilities of area chair by utilizing multiple models collaboratively, leading to improved decision-making. To verify this, we compare ReFeR's reasoning capabilities against other frameworks, including zero-shot-CoT, single-agent frameworks, and same-model multi-agent frameworks. Table \ref{tab:reasoning-results} presents the results of these experiments, with all results averaged across 3 runs, following the setup in \citep{chen2024reconcile}.
Key observations:
\begin{itemize}[noitemsep]
\item On average, ReFeR outperforms all other baselines across the tested benchmarks.
\item In the BBH Date Understanding benchmark, debating-type frameworks like Multi-Agent Debate show better results than ReFeR. This may be attributed to the nature of the benchmark, which involves understanding dates and resolving conflicts. Such tasks benefit from inter-agent discussions, which are possible in a debating setup but not in ReFeR's hierarchical framework.
\item ReFeR outperforms baselines on the AQuA benchmark because the hierarchical structure allows the area chair to synthesize peer inputs efficiently, avoiding confusion. In contrast, debate formats may cause models to introduce conflicting reasoning, which is less effective for tasks requiring precise reasoning like AQuA. 
\item Considering overall cost and performance, both variants of ReFeR demonstrate significant advantages in terms of cost-efficiency compared to corresponding multi-agent models.
\end{itemize}

\begin{table}[ht!]
\centering
\scriptsize
\caption{\textbf{Experimental results on Reasoning tasks.} Comparison of ReFeR performance (accuracy) with single-agent and multi-agent method baselines. All results are averaged across 3 runs. Cost*- Costs are shown as relative to the most expensive method.}
\label{tab:reasoning-results}
\begin{tabular}{|p{2cm}|l|c|c|c|c|c|c|}
\hline
\textbf{Method Type} & \textbf{Methods} & \textbf{AQuA} & \textbf{BBH\_DU} & \textbf{CSQA} & \textbf{GSM8k} & \textbf{Average} & \textbf{Cost*} \\
\hline
\multirow{3}{*}{Peer Agents} & Llama-3.1-8B & 26.3{\scriptsize\,$\pm$\,5.1} & 28.0{\scriptsize\,$\pm$\,7.8} & 68.3{\scriptsize\,$\pm$\,4.0} & 40.0{\scriptsize\,$\pm$\,11.4} & 40.7{\scriptsize\,$\pm$\,7.1} & 0.03 \\
 & Mistral Nemo-12B & 43.0{\scriptsize\,$\pm$\,3.6} & 55.7{\scriptsize\,$\pm$\,4.6} & 65.7{\scriptsize\,$\pm$\,6.1} & 54.7{\scriptsize\,$\pm$\,11.5} & 54.8{\scriptsize\,$\pm$\,6.5} & 0.05 \\
 & Gemma-2-9B & 50.7{\scriptsize\,$\pm$\,2.3} & 70.3{\scriptsize\,$\pm$\,6.5} & 75.7{\scriptsize\,$\pm$\,4.5} & 79.3{\scriptsize\,$\pm$\,4.0} & 69.0{\scriptsize\,$\pm$\,4.3} & 0.04 \\
\hline
\multirow{2}{*}{Single Agent} & zero-shot-CoT & 60.7{\scriptsize\,$\pm$\,1.5} & 88.0{\scriptsize\,$\pm$\,1.7} & 76.0{\scriptsize\,$\pm$\,1.7} & 95.3{\scriptsize\,$\pm$\,1.2} & 80.0{\scriptsize\,$\pm$\,1.5} & 0.06 \\
 & Self Correction & 78.7{\scriptsize\,$\pm$\,0.6} & \underline{92.7}{\scriptsize\,$\pm$\,0.6} & 54.0{\scriptsize\,$\pm$\,1.0} & 93.3{\scriptsize\,$\pm$\,1.5} & 79.7{\scriptsize\,$\pm$\,0.9} & 0.20 \\
\hline
\multirow{2}{*}{Multi-Agent} & Multi-Agent Debate & 54.0{\scriptsize\,$\pm$\,3.5} & \textbf{94.3}{\scriptsize\,$\pm$\,0.6} & \underline{79.0}{\scriptsize\,$\pm$\,1.7} & \underline{95.7}{\scriptsize\,$\pm$\,0.6} & 80.8{\scriptsize\,$\pm$\,1.6} & 0.97 \\
 & Multi-Agent Peer Review & 62.0{\scriptsize\,$\pm$\,4.4} & 91.7{\scriptsize\,$\pm$\,0.6} & 77.0{\scriptsize\,$\pm$\,1.0} & 95.3{\scriptsize\,$\pm$\,1.5} & 81.5{\scriptsize\,$\pm$\,1.9} & 1.0 \\
\hline
\multirow{2}{*}{Ours} & ReFeR Turbo & \textbf{85.0}{\scriptsize\,$\pm$\,1.0} & 92.0{\scriptsize\,$\pm$\,2.0} & \textbf{79.3}{\scriptsize\,$\pm$\,1.2} & \textbf{96.0}{\scriptsize\,$\pm$\,0.0} & \textbf{88.1}{\scriptsize\,$\pm$\,1.0} & 0.93 \\
 & ReFeR Lite & \underline{81.0}{\scriptsize\,$\pm$\,2.0} & 91.0{\scriptsize\,$\pm$\,2.0} & \textbf{79.3}{\scriptsize\,$\pm$\,1.2} & 93.3{\scriptsize\,$\pm$\,2.1} & \underline{86.2}{\scriptsize\,$\pm$\,1.8} & 0.18 \\
\hline
\end{tabular}
\end{table}

\section{Analysis}
\label{section: analysis}
In this section, we perform an analysis of the framework to understand the impact of different components and choices. 

\subsection{Peer Ablation}
\label{subsection: peer-ablation}
\begin{figure*}[!t]
\centering\includegraphics[width=3.5in]{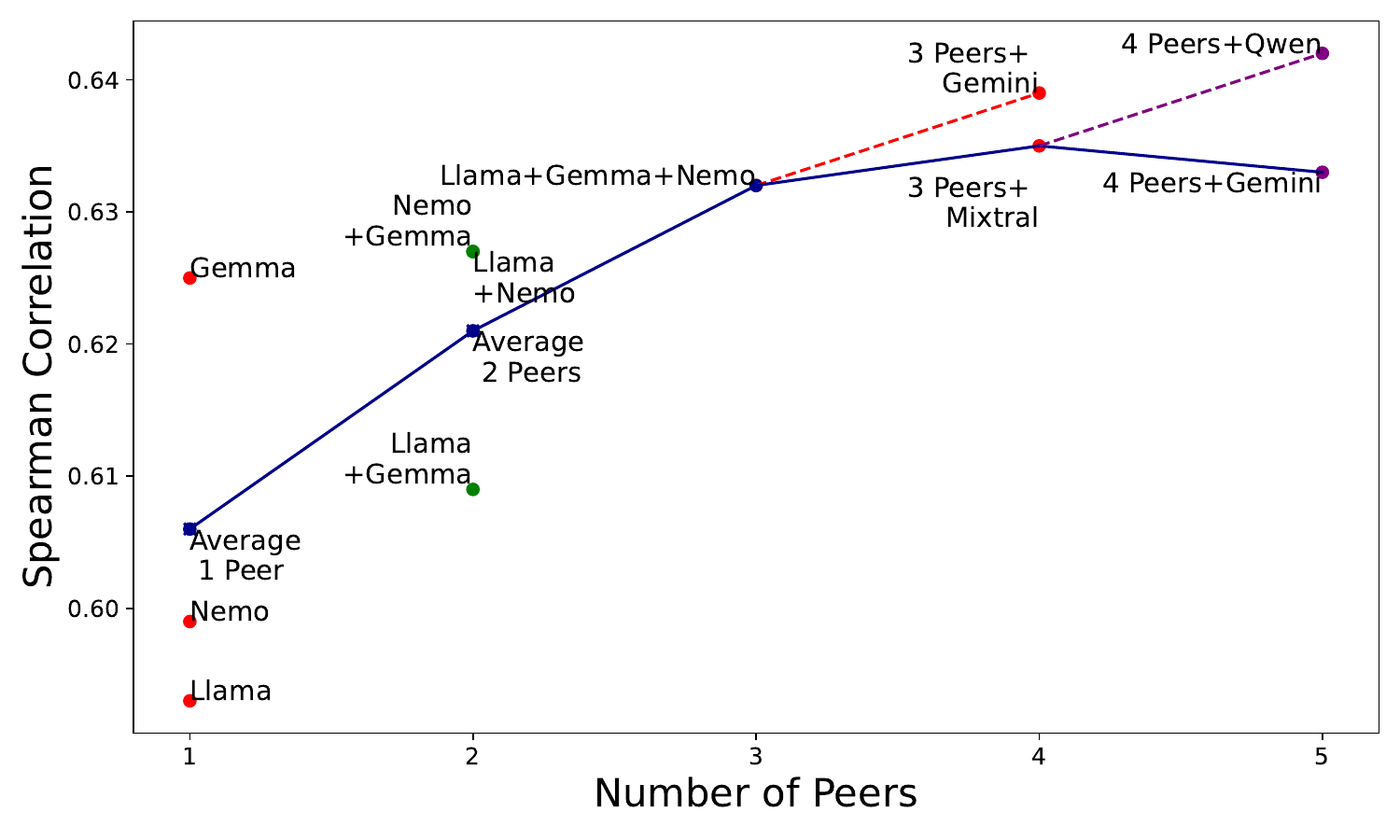}
\caption{\textbf{Framework Ablation.} Results obtained on ReFeR-Turbo by progressively adding different peers for the TopicalChat Dataset. The points in the figure indicate the performance of ReFeR when specific labelled peers were used in conjunction with the area chair (GPT-4o-mini). ``3 Peers'' refers to the Llama, Nemo, and Gemma models being used as peers. ``4 Peers'' includes the same 3 peers along with the Mixtral model added as the fourth peer. Detailed results are presented in Table \ref{tab:peer-ablation}.}
\label{fig:peer-ablation}
\end{figure*}

To evaluate the impact of number of peer agents and composition on ReFeR's performance, we conducted a peer ablation study using the TopicalChat dataset shown in Fig.~\ref{fig:peer-ablation}.

Our findings indicate that increasing the number of peers generally improves the framework's overall correlation, as evidenced by the main branch in Fig. \ref{fig:peer-ablation}. We experimented with varying peer combinations and numbers to distinguish between the effects of adding another peer versus a better-performing peer. Due to the impracticality of exploring all possible combinations with six peers, we selected a subset based on individual performances, costs, and model sizes.

Fig. \ref{fig:peer-ablation} demonstrates that while the framework's average performance generally increases with more peers, adding a relatively weaker model can result in performance better than the base (1 peer) but not necessarily the highest overall. For instance, with five peers, the combination of four peers plus Qwen yields the best performance, whereas four peers plus Gemini (weaker at this task) performs closer to the three-peer configuration.
Notably, the performance gain from four or five peers compared to three peers is not substantial. This observation suggests that using three peers may be an optimal choice, balancing performance improvements with computational efficiency.

\subsection{Selecting Peers and area chair}
\label{subsection: peer-selection}

For optimal peer selection, we recommend using a group of capable peers chosen based on their individual performances in performance assessment. After assessing individual performances, top-performing peers can be selected considering both cost and performance. As shown in Table \ref{tab:peer-ablation}, Gemma2-9B is the top performer across all metrics, while Mistral Nemo 12B and Llama-3.1 8B offer comparable performance at lower costs. Consequently, we selected Gemma2-9B, Llama-3.1 8B, and Mistral Nemo 12B as our peers. This selection is also crucial for enabling local GPU deployment of the peers.

To understand the framework's effectiveness under various conditions, we conducted a study by fixing the peers and changing the area chair. Particularly, we choose an area chair which is relatively weaker than not just GPT-4o-mini but also our best peer, Gemma2-9B, at this task. Hence, we choose Qwen1.5 - 72B. Table \ref{tab: peer-selection} presents the results using the ReFeR Lite setting on the TopicalChat dataset. We observed that although we used Qwen as AC (whose individual performance is less than the best peer), we get improved performance. This leads to a crucial observation that as long as AC is relatively stronger than some peers it can be used in the framework to get improved performance.

\begin{table}[ht!]
\centering
\scriptsize
\caption{\textbf{Results on TopicalChat using the open-source model Qwen1.5-72B as the area chair.} We were unable to include results for ReFeR Turbo with Qwen as the area chair due to the limitation of not being able to use n=20.}
\label{tab: peer-selection}
\begin{tabular}{|l|cc|cc|cc|cc|cc|}
\hline
\multirow{2}{*}{\textbf{Method}} & \multicolumn{2}{c|}{\textbf{Coherence}} & \multicolumn{2}{c|}{\textbf{Engagingness}} & \multicolumn{2}{c|}{\textbf{Groundedness}} & \multicolumn{2}{c|}{\textbf{Naturalness}} & \multicolumn{2}{c|}{\textbf{Avg}} \\
\cline{2-11}
 & $\rho$ & $\tau$ & $\rho$ & $\tau$ & $\rho$ & $\tau$ & $\rho$ & $\tau$ & $\rho$ & $\tau$ \\
\hline
(Peer) Llama-3.1-8B & 0.417 & 0.357 & 0.418 & 0.357 & 0.488 & 0.455 & 0.346 & 0.289 & 0.417 & 0.365 \\
(Peer) Mistral Nemo-12B & 0.416 & 0.352 & 0.567 & 0.475 & 0.453 & 0.424 & 0.396 & 0.339 & 0.458 & 0.397 \\
(Peer) Gemma-2-9B & \textbf{0.549} & \textbf{0.465} & \textbf{0.623} & \textbf{0.534} & \underline{0.583} & \underline{0.545} & \underline{0.520} & 0.431 & \textbf{0.568} & \textbf{0.494} \\
Qwen (Individual Performance) & 0.465 & 0.399 & 0.524 & 0.459 & 0.471 & 0.441 & 0.508 & \underline{0.437} & 0.492 & 0.434 \\
ReFeR Lite (Qwen) & \underline{0.496} & \underline{0.422} & \underline{0.609} & \underline{0.522} & \textbf{0.587} & \textbf{0.550} & \textbf{0.527} & \textbf{0.450} & \underline{0.555} & \underline{0.486} \\
\hline
\end{tabular}
\end{table}

This observation aligns with the original analogy of research paper peer review, where the area chair is typically a senior researcher with a potentially better understanding than most peer reviewers, thus being given more importance or final judgment authority. In cases where performance assessment is not feasible to determine the most suitable models, the LLM Leaderboard on 
 \citet{openllm_leaderboard} can be consulted to select appropriate models based on the specific task requirements, cost considerations, GPU availability, and time constraints.

\subsection{Error Analysis}
\label{subsection: error-analysis}
\begin{figure}[ht!]
    \centering
    \begin{subfigure}[b]{0.49\textwidth}  
        \centering
        \includegraphics[width=\textwidth]{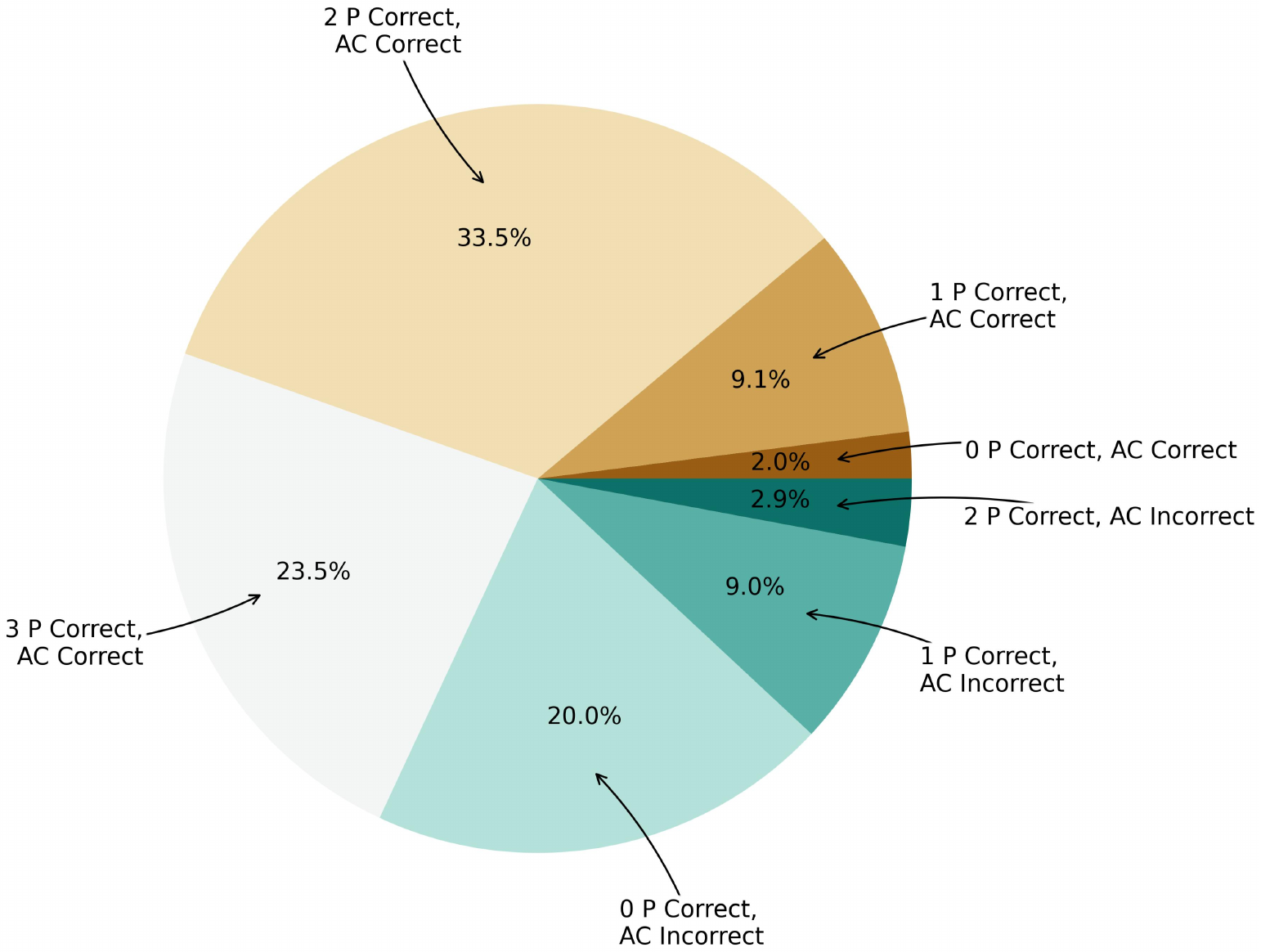}  
        \caption{Evaluation (TopicalChat)}  
        \label{fig:piechart-nlg}  
    \end{subfigure}
    \begin{subfigure}[b]{0.49\textwidth}  
        \centering
        \includegraphics[width=\textwidth]{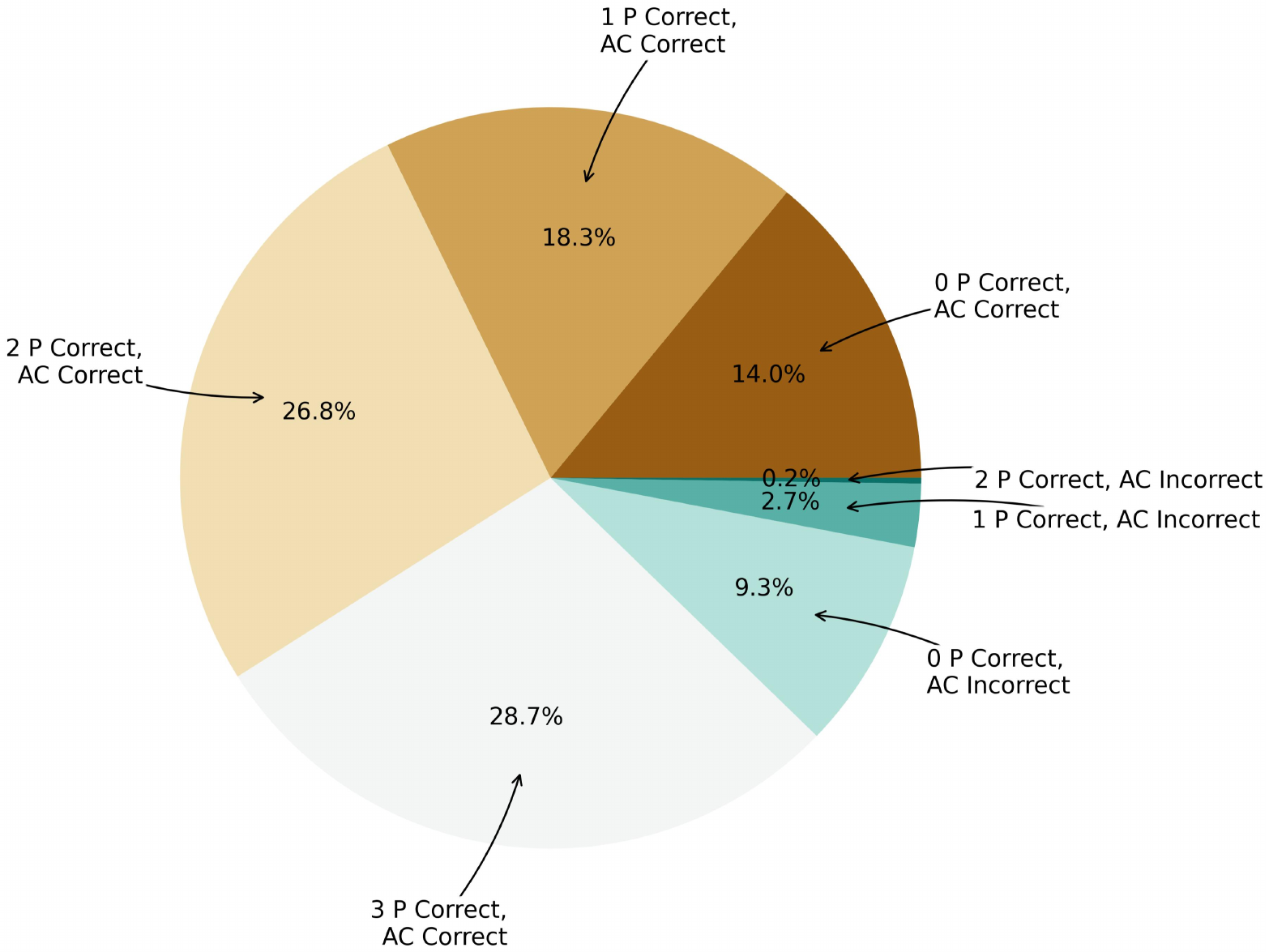} 
        \caption{Reasoning}  
        \label{fig:piechart-reasoning}  
    \end{subfigure}
    \caption{\textbf{Performance analysis wrt framework scale.} Pie-charts showing Peer and AC performance on evaluation and reasoning tasks. (P- Peer model, AC- area chair Model)} 
    \label{fig:piechart-error-analysis}
\end{figure}

To assess the framework's effectiveness in both evaluation and reasoning tasks, we conducted an error analysis, with results shown in Fig. \ref{fig:piechart-error-analysis}. In this analysis, a TopicalChat sample's evaluation score for each metric is considered correct if it falls within a given threshold range of 25\%. In the TopicalChat dataset evaluation, the area chair provided correct scores 42.6\% of the time when one or two peers provided a correct answer, demonstrating the AC's ability to leverage partially correct peer scores effectively. The AC made mistakes only 11.9\% of the time when at least one peer was correct. However, the AC was correct only in 2\% of the cases where all the peers were incorrect, suggesting that the AC may require at least one correct peer input to avoid confusion and give a correct score. For reasoning tasks, the AC was incorrect for only 2.9\% of cases where atleast one peer is correct, showing similiar observation as evaluation. And the AC was correct 14\% of the time, even when all peers were incorrect, indicating a better ability to disregard clearly incorrect answers from the smaller peer models. This suggests that the AC, when using reasoning tasks, may not always rely on peers and can function independently in such cases.

More analysis about prompt optimization, communication strategies, inference, and statistical significance tests can be found in Appendix \ref{appendix:prompt-optimize}, \ref{appendix:communication-strategies}, \ref{appendix: inference} and \ref{appendix: statistical-tests} respectively.
\section{Related Work}
\label{section: related-work}
\noindent\textbf{NLG \& Multimodal Evaluation:}
Recent advancements in NLG evaluation include GPTScore \citep{fu2023gptscore}, which uses generative pre-training models to assess text quality, and G-Eval \citep{liu2023geval}, employing a chain-of-thoughts approach with form-filling methodology. \citet{chiang2023closer} highlighted limitations in G-Eval's automated CoT alignment with human evaluations. TIGERScore \citep{jiang2023tigerscore} offers detailed error analysis using fine-tuned Llama-2, while FusionEval \citep{shu2024fusioneval} integrates auxiliary evaluators with a primary LLM for scoring. X-Eval \citep{liu2023xeval} introduces a two-stage instruction tuning framework for diverse evaluation dimensions. ChatEval \citep{chan2023chateval} proposes a multi-agent referee system using autonomous debating among agents with different personas to evaluate response quality. While similar to our approach, it primarily relies on debate methodology using the same models under varied personas, whereas our method employs diverse models as peers and area chairs with a richer prompting schema. In multimodal evaluation, CLIP Score \citep{clipscore}, Image-Reward \citep{xu2023imagerewardlearningevaluatinghuman}, and Pick Score \citep{kirstain2023pickapicopendatasetuser} assess image-text alignment using pre-trained models. Deep learning methods like HyperIQA \citep{Su_2020_CVPR} and IP-IQA \citep{qu2024bringingtextualpromptaigenerated} have shown improvements in this domain. Later, X-IQE \citep{chen2023xiqeexplainableimagequality} introduced using VLMs for the task of image-quality assessment.

\noindent\textbf{Reasoning using LLMs:}
Single-agent methods like Zero-shot CoT \citep{kojima2023largelanguagemodelszeroshot} have improved language models' reasoning capabilities using Chain-of-Thought prompting. Self-correction \citep{huang2024largelanguagemodelsselfcorrect} mimics human self-reflection to address reasoning errors. In multi-agent frameworks, \citet{du2023improvingfactualityreasoninglanguage} introduced a same-model approach using peer solutions for individual improvement, while \citet{pham2023let} proposed embedding-based communication to optimize reasoning. \citet{xu2023reasoninglargelanguagemodels} developed a framework inspired by academic peer review, emphasizing iterative improvement through peer feedback. This differs from our method, which involves an area chair reviewing peer responses without direct inter-peer communication. ReConcile \citep{chen2024reconcile} structured a multi-model, multi-agent framework as a round table conference, demonstrating enhanced reasoning through discussion and consensus. We expand on why we did not consider ReConcile as a baseline in the appendix \ref{appendix: reconcile}. \citet{wang2024unleashing} proposed selecting the most coherent response from multiple reasoning chains, offering an alternative approach to consensus-building and improving reasoning accuracy.

\section{Conclusion}
\label{section: conclusion}

In this work, we propose ReFeR (\textbf{Re}ason-\textbf{Fe}edback-\textbf{R}eview), a hierarchical model framework that utilizes smaller, capable models as peers and a powerful model as the area chair. The area chair leverages the reasoning and feedback from peers to provide a final review for evaluating given images or text. We demonstrate ReFeR's efficacy across two NLG evaluation tasks, two multimodal evaluation tasks, and four reasoning tasks, outperforming various baselines while maintaining performance and cost efficiency. We present two variants: ReFeR-Turbo and ReFeR-Lite. Notably, our Lite version achieves performance similar to other works and ReFeR-Turbo, while being significantly efficient. 

\section{Limitations}
\label{appendix: limitations}
Our framework, while robust in many aspects, has some limitations. One notable constraint is the potential computational cost when using large models as both peers and area chairs, especially in resource-limited environments. Additionally, the framework currently lacks an interactive discussion phase between peer models, which could further improve collective reasoning. In some scenarios, such as when a weaker model is used as the area chair, the performance may not be optimal. Lastly, while our framework has shown promising results on text and image evaluation tasks, it remains untested in other modalities, which could present unique challenges in scaling and complexity.

\section{Acknowledgments}
We would really like to thank Microsoft Research for their Accelerating Foundation Models Research initiative for generously funding our entire research and helping us with the GPUs and API credits. We would also like to thank Ganesh Jawahar for helping us with brainstorming the problem statement initially.

\section{Ethics Statement}
\label{appendix:ethics}

This work adheres to the ICLR Code of Ethics, ensuring that all evaluations and methodologies applied in the ReFeR framework were conducted with fairness, transparency, and integrity. Since ReFeR operates as a framework for evaluating machine-generated content, the primary ethical concerns are related to ensuring unbiased assessments and avoiding unintended model biases in evaluations. We carefully selected models to minimize potential biases, but the limitations of the models used could still introduce unintended biases, which we will continue to address in future improvements. No human subjects were involved in the experiments conducted for this study. Additionally, we commit to making our code and datasets available for further scrutiny and improvement.

\section{Reproducibility Statement}
\label{appendix:reproducibility}

To ensure the reproducibility of our results, we provide a detailed description of the ReFeR framework, including the structure of the hierarchical evaluation system and its variants. All hyperparameters, evaluation criteria, and the models used are described in the main text and appendices. The datasets utilized for evaluation and reasoning tasks are publicly available, as mentioned in Section 3.1. Additionally, we will release the source code, along with instructions for running the experiments, on an anonymous repository. Clear explanations for the model selection process, evaluation metrics, and experimental setups are also included to facilitate replication by other researchers.

\bibliography{iclr2025_conference}
\bibliographystyle{iclr2025_conference}

\appendix

\section{Future Works}
\label{appendix: future-works}
Future research can explore incorporating additional elements from the academic peer review process, such as the author discussion phase, to simulate a more interactive review environment. Expanding the framework to include evaluations beyond text and images, such as video and audio content, could further enhance its applicability. Another promising direction is to develop various communication strategies between peers and the area chair to optimize evaluation and feedback cycles. Moreover, experimenting with different numbers of area chairs of varying strength could help in understanding the impact of multiple, potentially conflicting, judgments on the final evaluation outcomes.

\begin{figure}[ht!]
    \centering
    \includegraphics[width=\columnwidth]{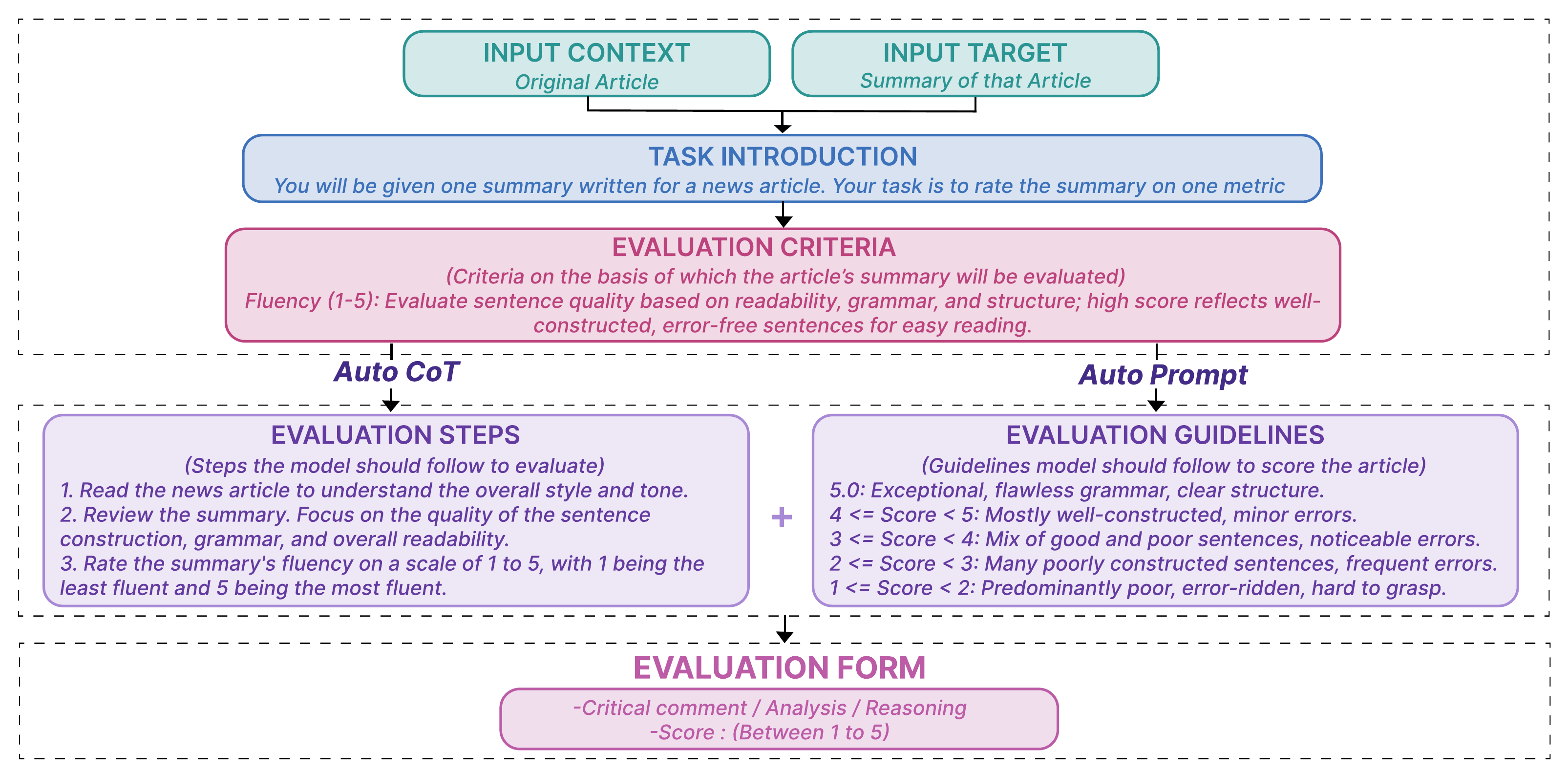}
    \caption{Prompting Schema}
    \label{fig:eval-schema}
\end{figure}

\begin{figure}[ht!]
    \centering
    \includegraphics[width=\columnwidth]{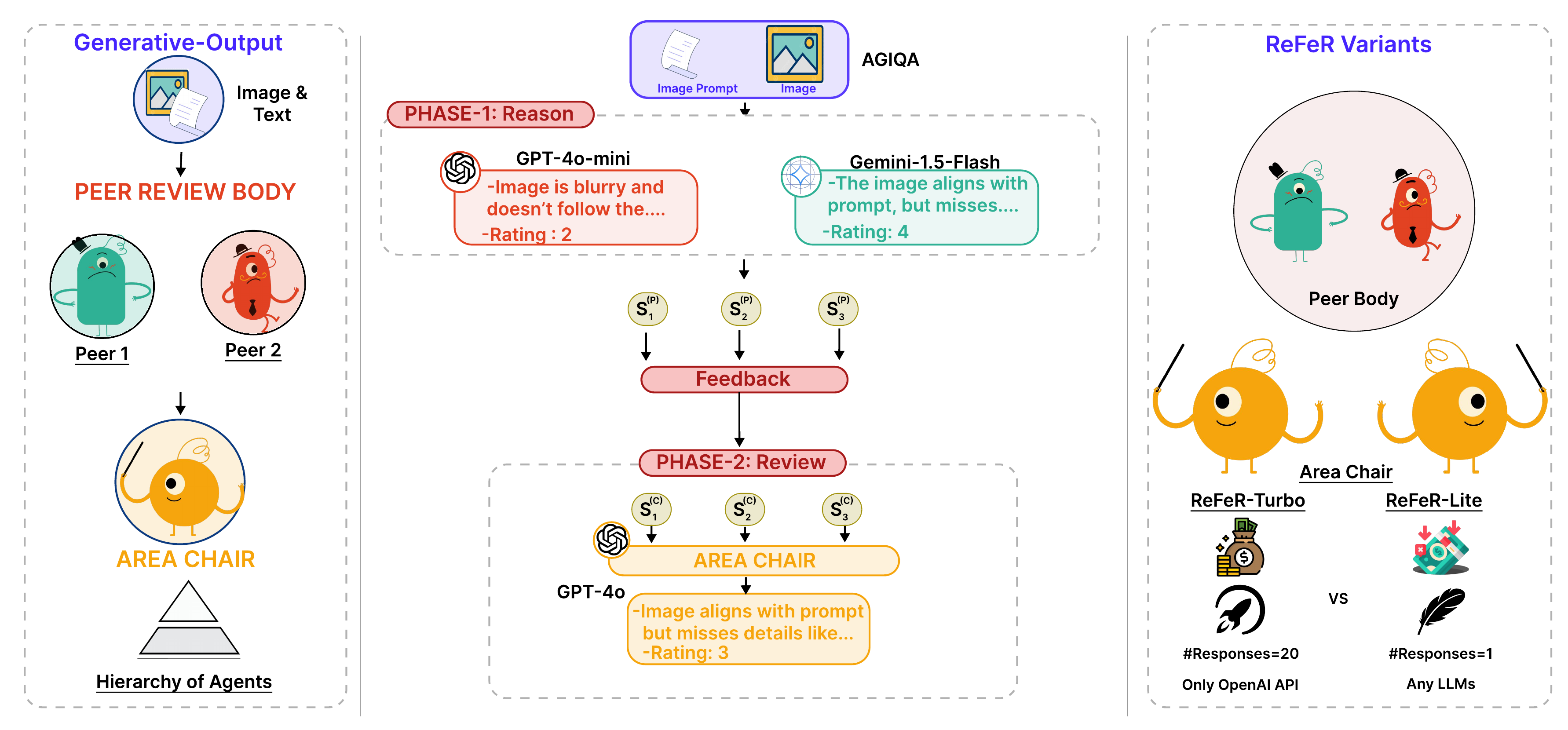}
    \caption{\textbf{Illustration of ReFeR for Multimodal evaluation shown on AGIQA dataset}. A similar version of ReFeR working on textual TopicalChat dataset is shown in \ref{fig:ReFeR}.}
    \label{fig:refer-multimodal}
\end{figure}

\section{ReFeR Algorithm}
\label{appendix: algorithm}

\begin{algorithm}[H]
\SetAlgoLined
\SetKwInOut{Input}{Input}
\SetKwInOut{Output}{Output}

\Input{
    \begin{itemize}[noitemsep]
        \item Generative output $O$ (text or image)
        \item Prompting schema $E_P$ for peers and $E_{AC}$ for area chair 
        \item Peer Models $P = \{P_1, P_2, P_3, \dots, P_K\}$ (K Peers) 
        \item Area Chair Model $AC$
        \item Variant $V \in \{\text{ReFeR-Turbo}, \text{ReFeR-Lite}\}$
        \item Number of responses $n$ (only for ReFeR-Turbo)
    \end{itemize}
}
\Output{
    \begin{itemize}
        \item Final Evaluation Score $S_{\text{final}}$
        \item Constructive Feedback $C_{\text{final}}$
    \end{itemize}
}

\BlankLine
\textbf{Phase 1: Peer Review Body Evaluation}\;
\ForEach{peer model $P_i \in P$}{
    \tcp{Each peer independently evaluates $O$ using prompting schema $E_P$}
    $(C_i, S_i) \gets \text{Evaluate}_{P_i}(G, E_P)$\; \label{line:peer_evaluation}
}

\BlankLine
\textbf{Phase 2: Area Chair Evaluation}\;

\If{$V = \text{ReFeR-Lite}$}{
    $n \gets 1$\;
}
\Else{
    $n \gets 20$\;
}
\For{$j = 1$ \KwTo $n$}{
    $(C_{\text{AC}}^{(j)}, S_{\text{AC}}^{(j)}) \gets \text{Evaluate}_{AC}^{(j)}(G, E_{AC}, \{(C_i, S_i)\}_{i=1}^K)$\; \label{line:ac_evaluation}
}
\tcp{Compute final score}
$S_{\text{final}} \gets \frac{1}{n} \sum_{j=1}^{n} S_{\text{AC}}^{(j)}$\;
$C_{\text{final}} \gets \text{Choose 1}(C_{\text{AC}}^{(1)}, \dots, C_{\text{AC}}^{(n)})$\;

\BlankLine

\Return{Final Evaluation Score $S_{\text{final}}$, Constructive Feedback $C_{\text{final}}$}\;

\caption{ReFeR Framework for Evaluating Generative Outputs}
\label{algo:refer}
\end{algorithm}

\textbf{Mathematical Notation Summary:}

\begin{multicols}{2}
\begin{itemize}[noitemsep, left=0em]
    \item $G$: Generative output to be evaluated.
    \item $E_P$: Prompt of peer.
    \item $E_{AC}$: Prompt of area chair.
    \item $P_i$: Peer agent $i$, for $i = 1, \dots, K$.
    \item $K$: Number of peer agents.
    \item $C_i$: Comment from peer agent $P_i$.
    \item $S_i$: Score from peer agent $P_i$.
\end{itemize}
\begin{itemize}[noitemsep, left=0em]
    \item $AC$: area chair agent.
    \item $n$: Number of independent evaluations by $AC$ in ReFeR-Turbo.
    \item $C_{\text{AC}}^{(j)}$, $S_{\text{AC}}^{(j)}$: Comment and score from the $j$-th evaluation by $AC$.
    \item $C_{\text{final}}$, $S_{\text{final}}$: Final comment and score.
\end{itemize}
\end{multicols}
In summary, the ReFeR framework formalizes the evaluation of generative outputs by modeling the process after the hierarchical peer review system, with mathematical rigor to facilitate clarity and reproducibility. This approach not only enhances the evaluation accuracy but also provides constructive feedback, aligning closely with human judgment and expectations in complex evaluation scenarios. 

\section{Hyperparameters}
\label{appendix:hyper-param}

For the ReFeR NLG Evaluation setup, following Analyze-Rate \citep{chiang2023closer}, we set these hyperparameters as follows, for the AreaChair GPT-4o-mini model- \textit{temperature=1, max\_tokens=256, top\_p=1, frequency\_penalty=0, presence\_penalty=0, stop=None, n=20} (varies for ReFeR Lite and Turbo). For the peer models, we use the default hyperparameters except for the \textit{max\_tokens=128}. For multimodal evaluation, we use the same setup for the AC, but for the peers, we increase the \textit{max\_tokens} from 128 to 192 tokens.
For reasoning tasks, we follow the NLG evaluation setup for the area chair, but we don't set any limit on the \textit{max\_tokens} hyperparameter.  For the peer models, we increase \textit{max\_tokens} to \textit{256} and set the hyper-parameters \textit{temperature=1, top\_p=1}.

\section{Prompt Optimization}
\label{appendix:prompt-optimize}

Prompt optimization methods utilizing LLMs, such as OPRO \citep{yang2024largelanguagemodelsoptimizers}, APE \citep{zhou2023largelanguagemodelshumanlevel}, and ProTeGi \citep{pryzant-etal-2023-automatic}, often employ text-gradient or feedback-based techniques to refine prompts. These methods typically involve providing a capable LLM with error examples and obtaining feedback, which serves as a text gradient to adjust the prompt for improved performance. Such approaches have demonstrated effectiveness for short questions/requests and relatively simple tasks.

Table \ref{tab:prompt-optimization} compares our prompt with an optimized prompt using ProTeGi on the TopicalChat dataset. We utilized the default ProTeGi settings with GPT-4o-mini as the optimizer LLM. Our findings indicate that prompt optimization is time-consuming and incurs higher costs than evaluation itself due to the iterative improvement process of prompt optimization over a test set, and that too with long inputs for complex tasks like this. Moreover, our prompts with the proposed structure yield better correlation than the optimized prompts from ProTeGi.

We attribute this outcome to the limitations of these methods when dealing with extensive inputs, such as conversation history and responses in TopicalChat, where entire dialogues are provided to evaluate and rate NLG output on various metrics. Unlike G-Eval, which only provides scores, methods like Analyze-Rate and ours improve scores based on generated analyses. In these cases, both the analyses and ratings are crucial for understanding errors, as the singular numerical rating value offers insufficient insight into prompt issues. Additionally, even when detailed analyses are provided in multiple error examples for the prompt optimization, the gradient-based approach may struggle with long contexts, making it challenging for the model to identify specific prompt deficiencies and provide useful feedback.

\begin{table}[ht!]
\centering
\scriptsize
\caption{\textbf{Analysis of Prompt Optimization.} Comparison of Average (across 4 metrics) results for different prompts on TopicalChat dataset.}
\label{tab:prompt-optimization}
\begin{tabular}{|c|c|c|c|c|}
\hline
\multirow{2}{*}{\textbf{Method}} & \multicolumn{2}{c|}{\textbf{ProTeGi}} & \multicolumn{2}{c|}{\textbf{Ours}} \\
\cline{2-5} 
 & $\rho$ & $\tau$ & $\rho$ & $\tau$ \\
\hline 
Llama-3.1-8B & 0.347 & 0.303 & 0.386 & 0.337 \\
Mistral-Nemo-12B & 0.387 & 0.336 & 0.464 & 0.402 \\
Gemma-2-9B & 0.511 & 0.444 & 0.563 & 0.489 \\
ReFeR Turbo & 0.625 & 0.511 & 0.628 & 0.513 \\
\hline
\end{tabular}
\end{table}

The table shows the results of average results across 4 metrics for peers and the framework for Prompt optimized by ProTeGi vs Our Prompt generated through our prompting schema. We can see that the peers' performance declines with the optimized prompts. These prompts were the best prompts after 3 rounds of Prompt-Optimization with ProTeGi. But still their performance falls short to our prompting schema. And even though the framework is relatively close, it would still make a point on how the effort and costs for the prompt optimizations would not be worth it. Running ProTeGi prompt optimizations alone for peers and area chair costs $\sim4$ times the cost of evaluating using ReFeR-Turbo. 

\subsubsection*{Prompt Ablation}

\begin{table}[h!]
\centering
\scriptsize
\caption{\textbf{Analysis of Prompt Ablation.} Average results (across 4 metrics) on TopicalChat Dataset of ReFeR-Turbo with different Prompt structure.}
\label{tab:prompt-ablation}
\begin{tabular}{|c|c|c|c|}
\hline
 & \multirow{2}{*}{\textbf{Prompt Schema}} & \multicolumn{2}{c|}{\textbf{Average}}\\
\cline{3-4} 
 & & $\rho$ & $\tau$\\
\hline 
\multirow{2}{*}{\textbf{ReFeR-Turbo}} & G-Eval & 0.568 & 0.454 \\
& Analyze-Rate & 0.592 & 0.510\\
& Ours & 0.628 & 0.513\\
\hline
\end{tabular}
\end{table}

Table \ref{tab:prompt-ablation} shows the ReFeR framework performance on TopicalChat Dataset with different prompts. We can see from the results that our prompting Schema gives the highest performance with Analyze-rate being the second. From the significant difference, we can see that G-eval scores only prompting doesn't work for our ReFeR framework.

\section{Auto Prompt}
\label{appendix:Auto-prompt}

\subsection*{Example Implementation of Autoprompt:}

\textbf{Input AutoPrompt:}

\begin{mdframed}[style=MyFrame, frametitle={AutoPrompt for Engagingness Evaluation for TopicalChat}, backgroundcolor=white]

\textit{You are tasked with creating a clear and concise prompt for a task based on the provided prompt structure and examples from dataset. The prompt should be written in such a way that it can be easily understood and followed by another LLM or human user performing the task. Your prompt should include the following:}

\begin{itemize}[left=0pt, noitemsep]
    \item \textit{A brief overview of the task.}
    \item \textit{Evaluation criteria explaining what metric the evaluation is going to be on.}
    \item \textit{Clear instructions for how to approach the task or evaluation steps.}
    \item \textit{Use the examples of dataset, analyze and understand how it is evaluated for the given metric and give a detailed Evaluation guidelines which will tell when to give a particular score.}
\end{itemize}

\textbf{\textit{Example Prompt Structure:}}

\textcolor{blue}{
\textit{You will be presented with a conversation between two individuals and given a potential response for the next turn in the conversation, along with a fact that the response is based on.}}

\textcolor{blue}{\textit{Your task is to evaluate the response on a single metric: Engagingness. The rating must be given after giving the analysis too.}}

\textcolor{blue}{\textit{Evaluation Criteria:}}

\textcolor{blue}{\textit{Engagingness (1-3): Assess whether the response is dull, moderately interesting, or highly engaging.}
}

\textcolor{blue}{\textit{Evaluation Steps:}}
\textit{\begin{enumerate}
    \item \textcolor{blue}{Read the conversation, the corresponding fact and the response carefully.}
    \item \textcolor{blue}{Rate the response on a scale of 1-3 for engagingness, according to the criteria above.}
\end{enumerate}}

\textcolor{blue}{\textit{Please ensure the prompt explains the rating scale from 1 to 3 clearly.
}\\}

\textbf{\textit{Examples for Task:}}

\textit{Example 1:}\\

\textcolor{blue}{\textit{Conversation History:}}\\
\textcolor{blue}{\textit{Person A: "I heard that flamingos get their pink color from their diet."\\
Person B: "Really? I didn’t know that! What do they eat to turn pink?"}}

\textcolor{blue}{\textit{Fact:} \textit{Flamingos get their pink color from consuming shrimp and algae rich in carotenoids.}}

\textcolor{blue}{\textit{Response:} \textit{"Yes! They eat shrimp and algae, which contain pigments called carotenoids that give them their pink color."}}

\textcolor{blue}{\textit{Rating:} \textit{3}}\\

\textit{Example 2:}\\

\textcolor{blue}{\textit{Conversation History:}}\\
\textcolor{blue}{\textit{Person A: "I just rewatched \textit{Interstellar}. The way they depict time dilation near a black hole is incredible."\\
Person B: "Yeah, the science behind it is pretty solid. What did you think of the emotional storyline with Cooper and his daughter?"\\
Person A: "I thought it was powerful how they tied in the human element with such a complex scientific concept."\\
Person B: "Do you think the film’s portrayal of black holes was accurate?"}}

\textcolor{blue}{\textit{Fact:} \textit{\textit{Interstellar} consulted physicist Kip Thorne to ensure the black hole depiction was as scientifically accurate as possible, especially with the time dilation effects.}}

\textcolor{blue}{\textit{Response:} \textit{"It was fine."}}

\textcolor{blue}{\textit{Rating:} \textit{1}}

\end{mdframed}

The above prompt is an example of Auto Prompt for the Engagingness metric in TopicalChat dataset. The text marked in \textcolor{blue}{blue}, i.e., the example prompt structure and the examples are the inputs to be given by the user to get the evaluation guidelines to be used in our proposed prompting schema. In the example above we use the prompt structure from Analyze-Rate engagingness prompt. Hence by prompting a capable LLM with an Auto Prompt crafted for a task giving the prompt structure and examples, we can obtain a prompt following our prompting schema for that task, including evaluation guidelines.

\textbf{Output based on the above prompt:}

\begin{mdframed}[style=MyFrame, frametitle={Engagingness Peer Prompt}]

\textit{You will be given a conversation between two individuals. You will then be given one potential response for the next turn in the conversation. The response concerns an interesting fact, which will be provided as well.}

\textit{Your task is to rate the responses on one metric.}
\textit{
Please make sure you read and understand these instructions carefully. Please keep this document open while reviewing, and refer to it as needed.}

\textit{Evaluation Criteria:}
\textit{Engagingness (1-3): Assess whether the response is dull, moderately interesting, or highly engaging.\\}

\textit{Evaluation Steps:}
\textit{\begin{enumerate}
    \item Read the conversation, the corresponding fact and the response carefully.
    \item Rate the response on a scale of 1-3 for engagingness, according to the criteria above.
\end{enumerate}}

\textit{Evaluation Guidelines:}\\
\textit{- A score of 1 (dull) means that the response is generic and dull.\\
- A score of 2 (somewhat interesting) means the response is somewhat interesting and could engage you in the conversation (e.g., an opinion, thought).\\
- A score of 3 (interesting) means the response is very interesting or presents an interesting fact.\\
}

\end{mdframed}

\section{Peer/AC Ablation Results}
\label{appendix: peer-ablation}

Table \ref{tab:peer-ablation} shows the individual performances of the 6 open-source models we chose and then the ablation of an increasing number of peers and the ablation of adding different models. The models we used are Llama-3.1-8B, Mistral-Nemo-12B, Gemma-2-9B, Mixtral-8x7B, Gemini-1.5-Pro, Qwen-1.5-72B.

\begin{table}[ht!]
\centering
\scriptsize
\caption{\textbf{Ablation analysis of the effect of different models used.} Comparison of ReFeR Turbo results on TopicalChat with Different Peer Configurations. The method column shows what peers were used with the AreaChair (GPT-4o-mini). 4 Peers in the last rows denotes the 3 peers (Llama, Nemo, Gemma) and Mixtral. *These rows show the individual performance of the peers, not the framework's performance when the peer is used.}
\label{tab:peer-ablation}
\begin{tabular}{|l|l|cc|cc|cc|cc|cc|}
\hline
\textbf{} & \textbf{Peers Used} & \multicolumn{2}{c|}{\textbf{Coherence}} & \multicolumn{2}{c|}{\textbf{Engagingness}} & \multicolumn{2}{c|}{\textbf{Groundedness}} & \multicolumn{2}{c|}{\textbf{Naturalness}} & \multicolumn{2}{c|}{\textbf{Average}} \\
\cline{3-12}
& & $\rho$ & $\tau$ & $\rho$ & $\tau$ & $\rho$ & $\tau$ & $\rho$ & $\tau$ & $\rho$ & $\tau$ \\
\hline
\multirow{7}{*}{Individual results$^*$} & Llama & 0.417 & 0.357 & 0.418 & 0.357 & 0.488 & 0.455 & 0.346 & 0.289 & 0.417 & 0.365 \\
& Nemo & 0.416 & 0.352 & 0.567 & 0.475 & 0.453 & 0.424 & 0.396 & 0.339 & 0.458 & 0.397 \\
& Gemma & 0.549 & 0.465 & 0.623 & 0.534 & 0.583 & 0.545 & 0.520 & 0.431 & 0.568 & 0.494 \\
& Mixtral & 0.440 & 0.373 & 0.552 & 0.467 & 0.491 & 0.458 & 0.469 & 0.390 & 0.488 & 0.422 \\
& Gemini & 0.352 & 0.300 & 0.460 & 0.387 & 0.498 & 0.466 & 0.419 & 0.352 & 0.432 & 0.376 \\
& Qwen & 0.465 & 0.399 & 0.524 & 0.459 & 0.471 & 0.441 & 0.508 & 0.437 & 0.492 & 0.434 \\
\cline{2-12}
& Average & 0.440 & 0.374 & 0.524 & 0.446 & 0.497 & 0.465 & 0.443 & 0.373 & 0.476 & 0.415 \\
\hline
\multirow{4}{*}{1 Peer} & Llama & 0.542 & 0.423 & 0.603 & 0.479 & 0.628 & 0.556 & 0.599 & 0.460 & 0.593 & 0.479 \\
&Nemo & 0.558 & 0.440 & 0.684 & 0.548 & 0.617 & 0.555 & 0.539 & 0.414 & 0.599 & 0.489 \\
& Gemma & 0.564 & 0.448 & 0.680 & 0.552 & 0.635 & 0.578 & 0.622 & 0.481 & 0.625 & 0.515 \\
\cline{2-12}
 & Average & 0.555 & 0.437 & 0.656 & 0.526 & 0.626 & 0.563 & 0.587 & 0.452 & 0.606 & 0.494 \\
\hline
\multirow{3}{*}{2 Peers} & Llama+Gemma & 0.565 & 0.440 & 0.656 & 0.524 & 0.593 & 0.535 & 0.621 & 0.481 & 0.609 & 0.495 \\
& Llama+Nemo & 0.577 & 0.450 & 0.692 & 0.553 & 0.621 & 0.570 & 0.621 & 0.480 & 0.627 & 0.513 \\
& Nemo+Gemma & 0.567 & 0.443 & 0.685 & 0.547 & 0.622 & 0.573 & 0.632 & 0.490 & 0.627 & 0.513 \\
\cline{2-12}
 & Average & 0.570 & 0.444 & 0.677 & 0.541 & 0.612 & 0.559 & 0.624 & 0.484 & 0.621 & 0.507 \\
\hline
3 Peers & Llama+Gemma+Nemo & 0.589 & 0.458 & 0.689 & 0.550 & 0.623 & 0.574 & 0.626 & 0.486 & 0.632 & 0.517 \\
\hline
\multirow{2}{*}{4 Peers} & 3 Peers + Mixtral & 0.596 & 0.463 & 0.682 & 0.541 & 0.629 & 0.572 & 0.634 & 0.494 & 0.635 & 0.517 \\
 & 3 Peers + Gemini & 0.601 & 0.469 & 0.688 & 0.550 & 0.644 & 0.590 & 0.623 & 0.485 & 0.639 & 0.523 \\
\hline
\multirow{2}{*}{5 Peers} & 4 Peers + Gemini & 0.584 & 0.455 & 0.686 & 0.545 & 0.623 & 0.572 & 0.640 & 0.495 & 0.633 & 0.517 \\
& 4 Peers + Qwen & 0.601 & 0.467 & 0.682 & 0.545 & 0.646 & 0.592 & 0.637 & 0.498 & 0.642 & 0.526 \\
\hline
\end{tabular}
\end{table}

\section{SummEval Results}
\label{appendix:SummEval}

\begin{figure}[ht]
    \centering
    \begin{subfigure}[t]{0.49\textwidth}
        \centering
        \includegraphics[width=\textwidth]{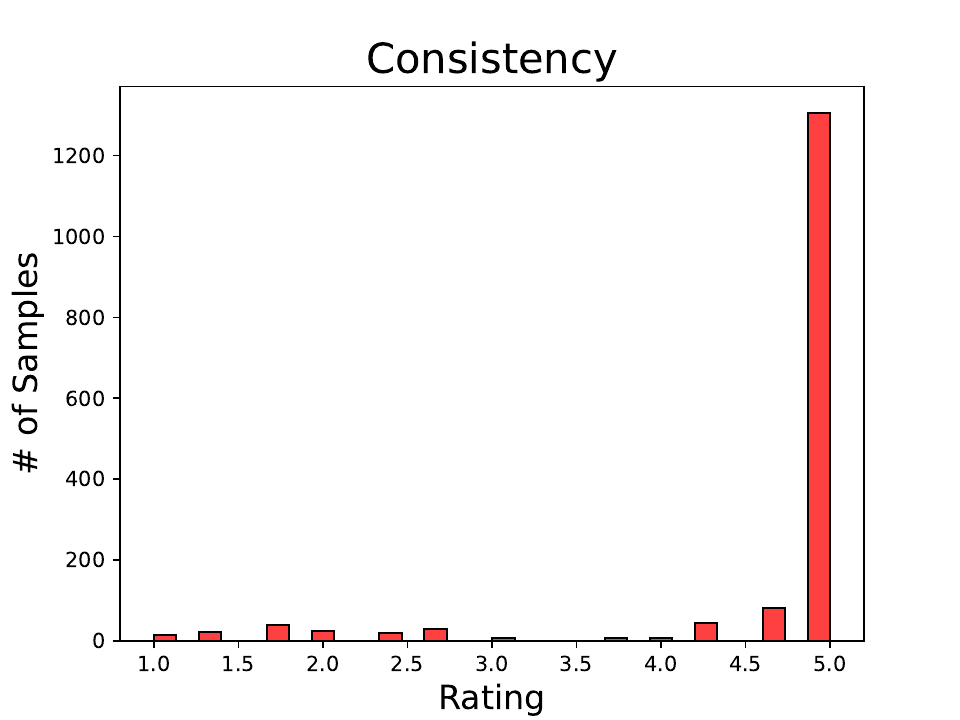}
        \caption{Consistency}
        \label{fig:sumemval-consistency}
    \end{subfigure}
    \hfill 
    \begin{subfigure}[t]{0.49\textwidth}
        \centering
        \includegraphics[width=\textwidth]{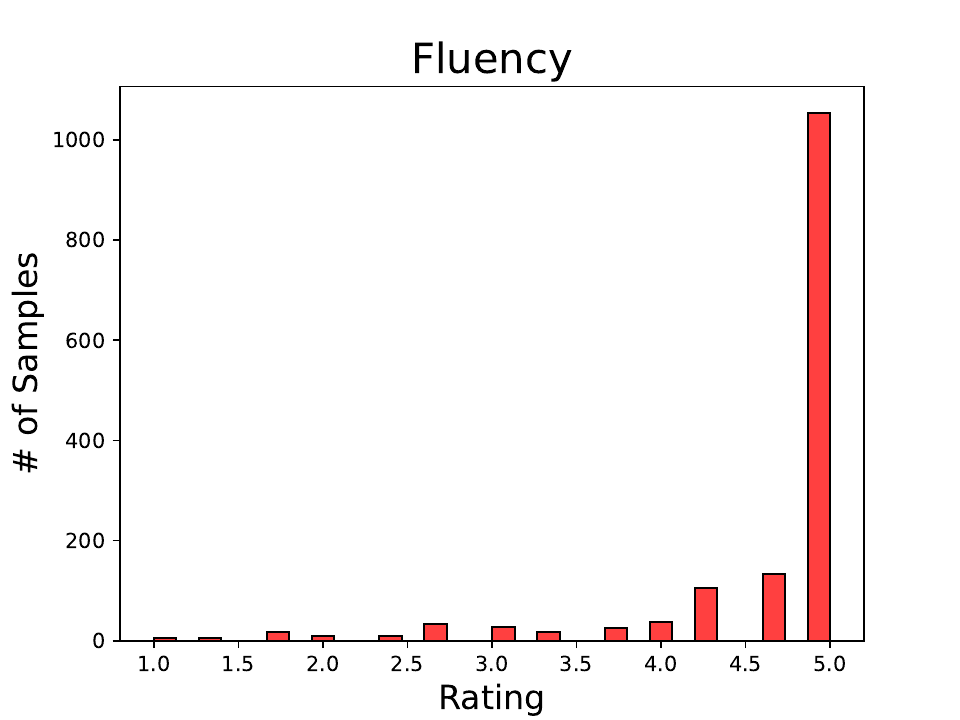}
        \caption{Fluency}
        \label{fig:summeval-fluency}
    \end{subfigure}
    \caption{\textbf{Ratings Distribution.} We show the distribution of human annotations for Consistency, FLuency metrics in the SummEval dataset}
    \label{fig:summeval-distr}
\end{figure}

We test our framework on the SummEval Dataset, comparing it with the baselines G-Eval and Analyze-Rate. We first show our individual peer performances, then the baselines, and finally, the two variants of our ReFeR framework. Before delving deep into the results of this benchmark, it is important to discuss the dataset distribution of SummEval. As shown in Fig.~\ref{fig:summeval-distr}, the dataset is highly skewed for the consistency and fluency metrics, with almost 1300+ and 1100+ samples having a score of 5 for consistency and fluency, respectively.

\begin{table}[ht!]
\centering
\scriptsize
\caption{\textbf{Performance analysis on SummEval dataset.} Comparison of various methods for NLG evaluation on SummEval.}

\label{tab:NLG-Comparison}
\begin{tabular}{|c|c|c|c|c|c|c|c|c|c|c|c|}
\hline
& \multirow{2}{*}{\textbf{Method}} & \multicolumn{2}{c|}{\textbf{Coherence}} & \multicolumn{2}{c|}{\textbf{Consistency}} & \multicolumn{2}{c|}{\textbf{Fluency}} & \multicolumn{2}{c|}{\textbf{Relevance}} & \multicolumn{2}{c|}{\textbf{Average}} \\
\cline{3-12}
& & $\rho$ & $\tau$ & $\rho$ & $\tau$ & $\rho$ & $\tau$ & $\rho$ & $\tau$ & $\rho$ & $\tau$ \\
\hline
\multirow{3}{*}{Peer Agents} & Llama-3.1-8B & 0.351 & 0.287 & 0.425 & 0.381 & 0.307 & 0.277 & 0.361 & 0.295 & 0.361 & 0.310 \\
& Mistral Nemo-12B & 0.367 & 0.296 & 0.383 & 0.340 & 0.239 & 0.211 & 0.368 & 0.303 & 0.339 & 0.287 \\
& Gemma-2-9B & \textbf{0.560} & \textbf{0.460} & 0.474 & \textbf{0.433} & \underline{0.387} & \textbf{0.347} & 0.517 & \underline{0.422} & \underline{0.484} & \textbf{0.415} \\
\hline
\multirow{2}{*}{Baselines} & Analyze-Rate & 0.533 & 0.392 & 0.382 & 0.305 & 0.353 & 0.283 & 0.430 & 0.320 & 0.425 & 0.325 \\
& G-Eval & 0.509 & 0.387 & \underline{0.475} & 0.386 & 0.334 & 0.290 & \textbf{0.571} & \textbf{0.433} & 0.472 & 0.374 \\
\hline
\multirow{2}{*}{Ours} & ReFeR & \underline{0.528} & \underline{0.403} & \textbf{0.478} & 0.390 & \textbf{0.425} & \underline{0.342} & \underline{0.521} & 0.395 & \textbf{0.488} & 0.382 \\
& ReFeR Lite & 0.483 & 0.400 & 0.472 & \underline{0.420} & 0.360 & 0.324 & 0.472 & 0.397 & 0.447 & \underline{0.385} \\
\hline
\end{tabular}
\end{table}

The skewed distribution in the Summeval dataset creates an imbalance in correlation evaluations. Smaller models, such as Gemma-2-9B, which may lack sensitivity to subtle differences in the data (e.g., article and summary), often give uniformly high scores that mimic the skewed human annotations, resulting in higher correlations. However, this does not reflect the model's true ability to understand and follow instructions. In contrast, larger models like GPT-4o-mini, which adhere more strictly to evaluation guidelines, tend to generate more varied scores. This often leads to lower correlations due to the skewed nature of the human annotations. Additionally, in cases where the ReFeR framework provides consistent scores, the result is a high number of tied predictions, which further lowers Kendall’s tau coefficient due to the large number of tied pairs. This complicates the interpretation of performance for larger models and more advanced frameworks, as the lack of variability in the dataset hampers an accurate assessment of model effectiveness. Given these challenges, although SummEval is a popular benchmark dataset used for NLG evaluation, we do not consider Summeval to be an appropriate benchmark for testing our methods, unless a uniformly distributed sample can be extracted—a difficult task given the inherent skewness of the annotations. Hence, even though  ReFeR-Turbo outperforms other baselines on this dataset, we present these results only in the appendix.

\section{Datasets}
\label{appendix:datasets}

\subsubsection*{NLG Evaluation}

\begin{itemize}[left=0pt, nosep]
    \item \textbf{SummEval} \citep{fabbri2021summeval} provides human assessments on four critical dimensions of summarization quality:  Coherence, Consistency, Fluency and Relevance, utilizing the CNN/DailyMail dataset \citep{hermann2015teaching} as its foundation. 
    \item \textbf{TopicalChat} \citep{gopalakrishnan2019topical}
    is a dataset of conversations. We use the dataset created by \citet{mehri2020usrunsupervisedreferencefree} using the TopicalChat dataset in which they give a possible next response generated by a language model for a given conversation history, and the human annotation score of the response on five attributes: Coherence, Engagingness, Groundedness, Naturalness, and Understandability. We exclude Understandibility, following the previous works G-Eval, Analyze-Rate and Uni-Eval\footnote{Uni-Eval shows results on the 4 metrics and uses the Understandability metric for transfer experiment, hence only 4 dimensons are shown in the following works. Refer to \citet{zhong2022unifiedmultidimensionalevaluatortext} for more details.}
\end{itemize}

\subsubsection*{MultiModal Evaluation}

\begin{itemize}[left=0pt, nosep]
    \item \textbf{ICQD} (Image Caption Quality Dataset) \citep{icqd2019}  focuses on the task of Quality Estimation (QE) for image captions. We use the test dataset which provides human ratings (0/1) on quality. We scale these average ratings to a scale of 0-100 for our evaluation.
    \item \textbf{AGIQA} (AI Generated Image Quality Assessment) \citep{zhang2023perceptualqualityassessmentexploration} presents a AGI quality assessment database, AGIQA-1K, which consists of 1,080 AGIs generated from diffusion models. They provide MOS (Mean Opinion Score) in the range of 0-5. We have observed that the dataset is skewed around certain scores around 3-3.5. So to test on a subset which has variance of image quality ratings, we select 500 samples, such that the data more or less equally spread on the rating range (0-5).
\end{itemize}

\subsubsection*{Reasoning}

\begin{itemize}[left=0pt, nosep]
    \item \textbf{AQuA} (Algebra Question Answering) \citep{ling-etal-2017-program}  dataset is designed to assess a model's reasoning abilities in solving algebraic word problems. It consists of multiple-choice math questions, where the model must understand and compute the correct answer from several options.
    \item \textbf{BBH-DU} (Big Bench Hard Date Understanding) \citep{srivastava2023imitationgamequantifyingextrapolating} dataset is part of the BIG-Bench benchmark. It focuses on testing a model's ability to comprehend and reason about date-related information, such as calculating durations and interpreting dates.
    \item \textbf{CSQA} (CommonsenseQA) \citep{aggarwal-etal-2021-explanations}  dataset is designed to test a model's understanding of commonsense knowledge through multiple-choice questions. Each question requires reasoning over general world knowledge, with answer choices based on various plausible but nuanced options, testing the model's ability to pick the most commonsensical answer.
    \item \textbf{GSM8k} (Grade School Math 8K) \citep{cobbe2021trainingverifierssolvemath} dataset is a collection of 8,000 challenging grade-school-level math word problems. It is designed to test a model’s ability to perform multi-step arithmetic reasoning and solve math problems requiring logical thinking and numerical computation.
\end{itemize}

\section{Note on ReConcile}
\label{appendix: reconcile}

ReConcile \citep{chen2024reconcile} is another relevant multi-agent framework that utilizes different LLMs with similar capabilities to engage in discussions and reach consensus. However, we exclude ReConcile from our baselines because its use of 3 LLMs of similar capabilities and makes it an unfair comparison to our framework, which employs 3 smaller models as peers and 1 larger model as the area chair. Simulating ReConcile with our setup would require excluding one of the models, either from the peer group or the area chair, which would lead to an unbalanced debate. In particular, if we use 2 smaller models and a large model, the debate would be dominated by the larger model, resulting in biased outcomes. For these reasons, we do not include ReConcile as a direct baseline.

\section{Finetuning}
\label{appendix: finetuning}
Utilizing Analysis from larger LLMs (``Area Chair''), we enhance smaller LLMs through instruction-tuning, using a dataset crafted from comprehensive evaluations. We use the analysis feedback generated within the ReFeR framework, transforming it into a useful resource for instructional tuning. This fine-tuning significantly improves smaller models performance, enabling them to reach or surpass their larger counterparts in evaluation tasks. We use Mistral-7B, since it can be easily deployable on a small GPU and finetune. We used the instruction-tuning dataset (final output of Area Chair) of ReFeR framework as the training data.

\begin{table}[h!]
\centering
\scriptsize
\caption{\textbf{Improving smaller models via instruction-tuning.} Finetuning Results for Mistral-7B model on TopicalChat Dataset}
\begin{tabular}{|l|c|c|c|c|c|c|c|c|c|c|}
\hline
\multirow{2}{*}{\textbf{Model}} & \multicolumn{2}{|c}{\textbf{Coherence}} & \multicolumn{2}{|c}{\textbf{Engagingness}} & \multicolumn{2}{|c}{\textbf{Groundedness}} & \multicolumn{2}{|c}{\textbf{Naturalness}} & \multicolumn{2}{|c|}{\textbf{Avg}} \\
\cline{2-11}
 & $\rho$ & $\tau$ & $\rho$ & $\tau$ & $\rho$ & $\tau$ & $\rho$ & $\tau$ & $\rho$ & $\tau$ \\
\hline
Mistral-7B No Finetune & 0.124 & 0.102 & 0.167 & 0.134 & 0.078 & 0.069 & 0.100 & 0.081 & 0.117 & 0.096 \\
Mistral-7B Finetuned & 0.457 & 0.348 & 0.626 & 0.486 & 0.487 & 0.437 & 0.493 & 0.377 & 0.516 & 0.412 \\
\hline
\end{tabular}
\label{tab:finetuning-topical}
\end{table}

\section{Communication Strategies:}
\label{appendix:communication-strategies}

\begin{table}[ht!]
\centering
\scriptsize
\caption{\textbf{Communication Strategies.} Results on TopicalChat showing different generation and communication strategies for ReFeR-Turbo.}
\label{tab:communication-strategy}
\begin{tabular}{|l|cc|cc|cc|cc|cc|}
\hline
\textbf{Communication} & \multicolumn{2}{c|}{\textbf{Coherence}} & \multicolumn{2}{c|}{\textbf{Engagingness}} & \multicolumn{2}{c|}{\textbf{Groundedness}} & \multicolumn{2}{c|}{\textbf{Naturalness}} & \multicolumn{2}{c|}{\textbf{Avg}} \\
\hline
 Peer Feedback to AreaChair& $\rho$ & $\tau$ & $\rho$ & $\tau$ & $\rho$ & $\tau$ & $\rho$ & $\tau$ & $\rho$ & $\tau$ \\
\hline
Comment Only  & \textbf{0.602} & \textbf{0.471} & 0.635 & 0.502 & \textbf{0.661} & \textbf{0.590} & \underline{0.587} & \underline{0.454} & \underline{0.621} & \underline{0.504} \\
Score Only & \underline{0.585} & \underline{0.454} & \textbf{0.673} & \textbf{0.535} & \underline{0.628} & \underline{0.577} & \textbf{0.625} & \textbf{0.484} & \textbf{0.628} & \textbf{0.513} \\
Both Comment \& Score  & 0.580 & 0.453 & \underline{0.642} & \underline{0.512} & 0.605 & 0.545 & 0.555 & 0.427 & 0.596 & 0.484 \\
\hline
\end{tabular}
\end{table}

The type of feedback provided by peers to the area chair plays a crucial role in determining overall effectiveness. We explored three communication strategies: passing only scores, passing only comments, and passing both comments and scores. Table \ref{tab:communication-strategy} presents the impact of different feedback strategies on the framework's performance. The results indicate that passing only scores to the AC yields the best performance, with passing only comments being a close second. This is likely because when both comments and scores are passed, the AC model becomes more prone to confusion due to conflicting analyses or scores, and the longer prompt inputs negatively affect its decision-making \citet{liu2023lostmiddlelanguagemodels}. Based on these findings, we adopt the scores-only strategy for all subsequent experiments with our framework.

\section{Inference}
\label{appendix: inference}

\begin{figure}[ht!]
    \centering
    \includegraphics[width=3in]{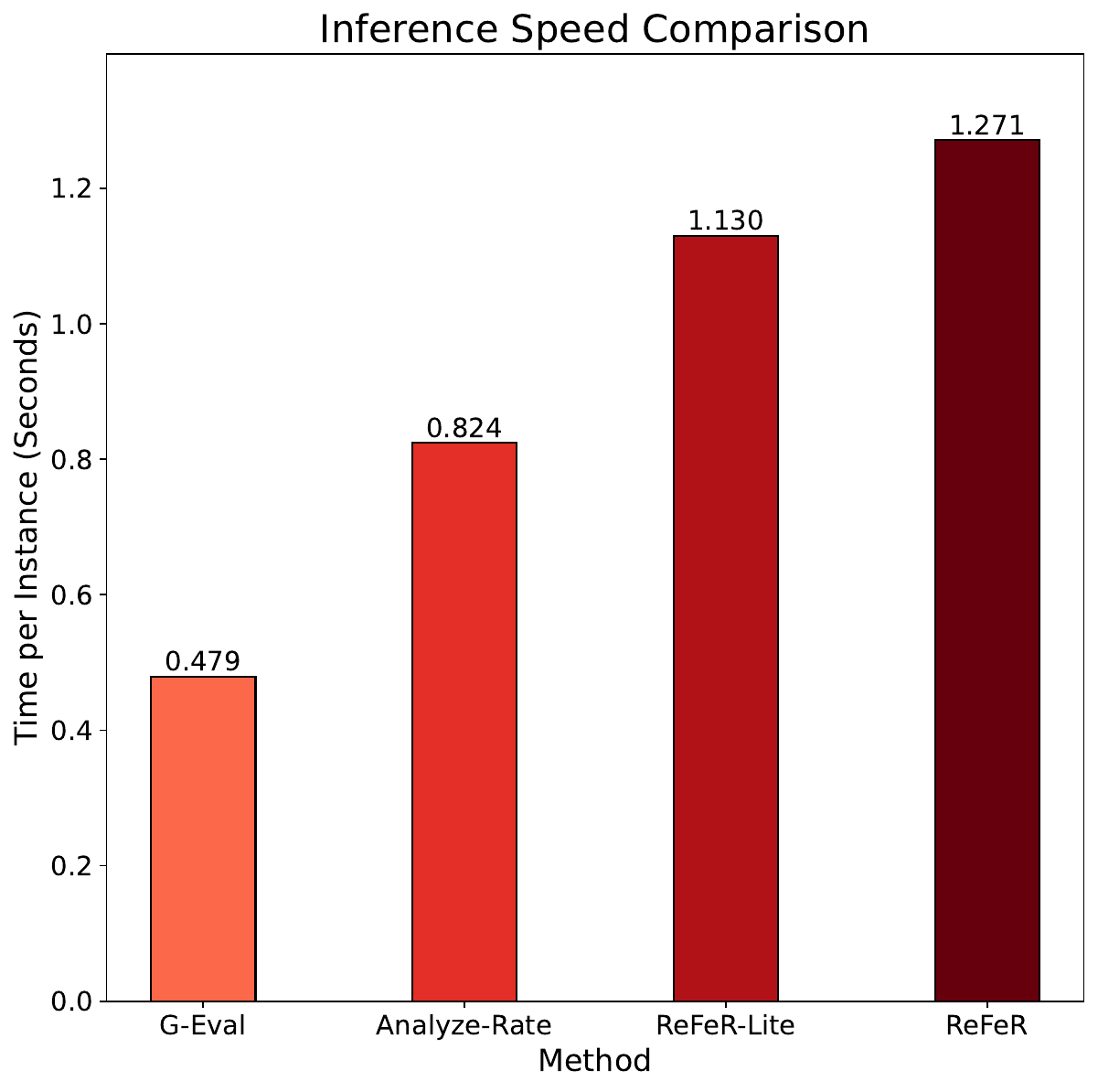}
    \caption{\textbf{Inference speed comparison with baselines}}
    \label{fig:inference-speed}
\end{figure}

Fig. \ref{fig:inference-speed} presents the time taken per instance for ReFeR Variants and baseline models. G-Eval demonstrates the fastest inference speed, as it only generates scores. In contrast, Analyze-Rate takes nearly twice as long as G-Eval, since it produces both an analysis and a rating. ReFeR-Lite and ReFeR-Turbo require only approximately 1.5 times the duration of Analyze-Rate, despite being multi-model frameworks. Notably, there is minimal difference between the Lite and Turbo variants due to the influence of the \textit{n} hyperparameter, indicating that the bulk of the processing time arises from the involvement of multiple models in the framework.

\section{Statistical Tests}
\label{appendix: statistical-tests}
\begin{table}[ht!]
\centering
\scriptsize
\caption{\textbf{Statistical Tests.} p-values for statistical tests between ReFeR Turbo and baselines on TopicalChat.}
\label{tab:stat-tests}
\begin{tabular}{|l|c|c|}
\hline
\textbf{Metric} & \textbf{ReFeR vs. Analyze rate} & \textbf{ReFeR vs. G-Eval} \\
\hline
Coherence & $2.34 \times 10^{-6}$ & $4.29 \times 10^{-13}$ \\
Engagingness & $3.70 \times 10^{-5}$ & $1.01 \times 10^{-17}$ \\
Groundedness & $1.15 \times 10^{-7}$ & $6.31 \times 10^{-12}$ \\
Naturalness & $1.19 \times 10^{-4}$ & $0.0736$ \\
\hline
\end{tabular}
\end{table}

We conducted the paired t-test to compare our method with the baseline and we report the p-values. As we can see for only one metric in G-Eval, the p-value is greater than the typical significance level (0.05). Hence, from the p-values, we can see that ReFeR is, in general, statistically significantly better than G-Eval and Analyze Rate.

\section{Prompts}
\label{appendix:Prompts}

\subsection{NLG Evaluation}

\subsubsection*{TopicalChat}

\begin{mdframed}[style=MyFrame, frametitle={Coherence Peer Prompt}]
\textit{You will receive a dialogue between two people. Following that, there will be one suggested reply for the next part of the conversation, along with a related interesting fact.}

\textit{Your job is to assess how coherent the suggested reply is, focusing on its ability to seamlessly continue the dialogue while also considering the overall context of the conversation, including the provided fact.}

\textit{Please read and understand these instructions carefully. You may refer back to them as needed.}

\textit{Assessment Criteria:}

\textit{Coherence (1-3): How well does the response continue the conversation?\\}
\textit{- A score of 1 (no) indicates that the reply significantly shifts the topic or disregards the ongoing conversation entirely.\\
- A score of 2 (somewhat) suggests that the response makes a vague reference to the conversation but fails to effectively engage with the dialogue or the accompanying fact.\\
- A score of 3 (yes) signifies that the response stays on topic, acknowledges the previous dialogue, and draws a clear and relevant connection to the interesting fact provided while maintaining the overall conversational flow.
}
\\

\textit{Assessment Process:}
\textit{\begin{enumerate}
    \item Review the conversation history for context and flow, focusing on how well the suggested reply relates to the previous exchanges.
    \item Examine the suggested reply for its relevance and engagement with the ongoing dialogue.
    \item Consider how well the reply connects with the interesting fact while also evaluating its contribution to the conversation as a whole.
    \item Assign a coherence score of 1, 2, or 3, taking into account both the conversational progression and the connection to the fact.
\end{enumerate}}

\textit{Example: }

\textit{Conversation History: \{\{Conversation\}\}}

\textit{Corresponding Fact: \{\{Contextual Fact\}\}}

\textit{Response: \{\{Generated Response\}\}}

\textit{Evaluation Form (Answer by starting with ``Analysis:'' to analyze the given example regarding the evaluation criteria as concise as possible, and then give the numeric rating on the next line by ``Rating''.)}

\textit{Coherence:}

\end{mdframed}

\begin{mdframed}[style=MyFrame, frametitle={Coherence AreaChair Prompt}]
\textit{Navigate through a simulated conversation between two individuals, followed by a provided potential response incorporating an intriguing fact. Your role is to assess the responses based on the coherence metric.}

\textit{Alongside your evaluation, you will also receive initial evaluations from three large language models, referred to as the assistants' evaluations. Please read the instructions and criteria below carefully and use them as a guide in your evaluation, critically assessing the conversation, and the assistants' inputs.}

\textit{Ensure a meticulous understanding of the instructions. Keep this document accessible for reference during the evaluation.}

\textit{Evaluation Criteria:}

\textit{Coherence (1-3): Assess whether the response seamlessly continues the conversation history.\\}
\textit{- A score of 1 (no) denotes a significant shift in topic or disregard for the conversation history.\\
- A score of 2 (somewhat) indicates a response with limited reference to the conversation history and a noticeable shift in topic.\\
- A score of 3 (yes) signifies an on-topic response that strongly acknowledges and builds upon the conversation history.\\
}

\textit{Evaluation Steps:}
\textit{\begin{enumerate}
    \item Thoroughly read the conversation history.
    \item Examine the potential response.
    \item Evaluate coherence based on the conversation history.
    \item Assign a coherence score of 1, 2, or 3.
\end{enumerate}}

\textit{Example: }

\textit{Conversation History: \{\{Conversation\}\}}

\textit{Corresponding Fact: \{\{Contextual Fact\}\}}

\textit{Response: \{\{Generated Response\}\}}

\textit{First Assistant's Evaluation: \{\{Peer\_response1\}\}}

\textit{Second Assistant's Evaluation: \{\{Peer\_response2\}\}}

\textit{Third Assistant's Evaluation: \{\{Peer\_response3\}\}}

\textit{Evaluation Form (Answer by starting with ``Analysis:'' to analyze the given example regarding the evaluation criteria as concise as possible, and then give the numeric rating on the next line by ``Rating''.)}

\textit{Coherence:}
\end{mdframed}

\begin{mdframed}[style=MyFrame, frametitle={Engagingness Peer Prompt}]

\textit{You will be given a conversation between two individuals. You will then be given one potential response for the next turn in the conversation. The response concerns an interesting fact, which will be provided as well.}

\textit{Your task is to rate the responses on one metric.}
\textit{
Please make sure you read and understand these instructions carefully. Please keep this document open while reviewing, and refer to it as needed.}

\textit{Evaluation Criteria:}

\textit{Engagingness (1-3): Is the response dull/interesting?\\}
\textit{- A score of 1 (dull) means that the response is generic and dull.\\
- A score of 2 (somewhat interesting) means the response is somewhat interesting and could engage you in the conversation (e.g., an opinion, thought).\\
- A score of 3 (interesting) means the response is very interesting or presents an interesting fact.\\
}

\textit{Evaluation Steps:}
\textit{\begin{enumerate}
    \item Read the conversation, the corresponding fact and the response carefully.
    \item Rate the response on a scale of 1-3 for engagingness, according to the criteria above.
\end{enumerate}}

\textit{Example: }

\textit{Conversation History: \{\{Conversation\}\}}

\textit{Corresponding Fact: \{\{Contextual Fact\}\}}

\textit{Response: \{\{Generated Response\}\}}

\textit{Evaluation Form (Answer by starting with ``Analysis:'' to analyze the given example regarding the evaluation criteria as concise as possible, and then give the numeric rating on the next line by ``Rating''.)}

\textit{Engagingness:}

\end{mdframed}

\begin{mdframed}[style=MyFrame, frametitle={Engagingness AreaChair Prompt}]
\textit{Navigate through a simulated conversation between two individuals, followed by a provided potential response incorporating an intriguing fact. Your role is to assess the responses based on the engagingness metric.}

\textit{Alongside your evaluation, you will also receive initial evaluations from three large language models, referred to as the assistants' evaluations. Please read the instructions and criteria below carefully and use them as a guide in your evaluation, critically assessing the conversation, and the assistants' inputs.}

\textit{Ensure a meticulous understanding of the instructions. Keep this document accessible for reference during the evaluation.}

\textit{Evaluation Criteria:}

\textit{Engagingness (1-3): Is the response dull or interesting?\\}
\textit{- A score of 1 (dull) means that the response is generic and uninteresting.\\
- A score of 2 (somewhat interesting) means the response is somewhat engaging and could capture interest (e.g., an opinion or thought).\\
- A score of 3 (interesting) means the response is highly engaging or presents an intriguing fact.\\
}

\textit{Evaluation Steps:}
\textit{\begin{enumerate}
    \item Read the conversation, the corresponding fact, and the response carefully.
    \item Rate the response on a scale of 1-3 for engagingness, according to the criteria above.
\end{enumerate}}

\textit{Example: }

\textit{Conversation History: \{\{Conversation\}\}}

\textit{Corresponding Fact: \{\{Contextual Fact\}\}}

\textit{Response: \{\{Generated Response\}\}}

\textit{First Assistant's Evaluation: \{\{Peer\_response1\}\}}

\textit{Second Assistant's Evaluation: \{\{Peer\_response2\}\}}

\textit{Third Assistant's Evaluation: \{\{Peer\_response3\}\}}

\textit{Evaluation Form (Answer by starting with ``Analysis:'' to analyze the given example regarding the evaluation criteria as concisely as possible, and then give the numeric rating on the next line by ``Rating''.)}

\textit{Engagingness:}
\end{mdframed}

\begin{mdframed}[style=MyFrame, frametitle={Groundedness Peer Prompt}]

\textit{You will be given a conversation between two individuals. You will then be given one potential response for the next turn in the conversation. The response concerns an interesting fact, which will be provided as well.}

\textit{Your task is to rate the responses on one metric.}
\textit{
Please make sure you read and understand these instructions carefully. Please keep this document open while reviewing, and refer to it as needed.}

\textit{Evaluation Criteria:}

\textit{Groundedness (0-1) given the fact that this response is conditioned on, determine whether this response uses that fact.\\}
\textit{- A score of 0 (no) means the response does not mention or refer to the fact at all.\\
- A score of 1 (yes) means the response uses the fact well.\\
}

\textit{Evaluation Steps:}
\textit{\begin{enumerate}
    \item Read the conversation between the two individuals.
    \item Identify the fact that is provided for the potential response.
    \item Read the potential response.
    \item Determine if the potential response uses or mentions the fact.
    \item Assign a score of 0 or 1 for groundedness based on whether the response uses the fact.
\end{enumerate}}

\textit{Example: }

\textit{Conversation History: \{\{Conversation\}\}}

\textit{Corresponding Fact: \{\{Contextual Fact\}\}}

\textit{Response: \{\{Generated Response\}\}}

\textit{Evaluation Form (Answer by starting with ``Analysis:'' to analyze the given example regarding the evaluation criteria as concise as possible, and then give the numeric rating on the next line by ``Rating''.)}

\textit{Groundedness:}

\end{mdframed}

\begin{mdframed}[style=MyFrame, frametitle={Groundedness AreaChair Prompt}]
\textit{Navigate through a simulated conversation between two individuals, followed by a provided potential response incorporating an intriguing fact. Your role is to assess the responses based on the groundedness metric.}

\textit{Alongside your evaluation, you will also receive initial evaluations from three large language models, referred to as the assistants' evaluations. Please read the instructions and criteria below carefully and use them as a guide in your evaluation, critically assessing the conversation, and the assistants' inputs.}

\textit{Ensure a meticulous understanding of the instructions. Keep this document accessible for reference during the evaluation.}

\textit{Evaluation Criteria:}

\textit{Groundedness (0-1): Given the fact that this response is conditioned on, determine whether this response uses that fact.\\}
\textit{- A score of 0 (no) means the response does not mention or refer to the fact at all.\\
- A score of 1 (yes) means the response uses the fact well.\\
}

\textit{Evaluation Steps:}
\textit{\begin{enumerate}
    \item Read the conversation between the two individuals.
    \item Identify the fact that is provided for the potential response.
    \item Read the potential response.
    \item Determine if the potential response uses or mentions the fact.
    \item Assign a score of 0 or 1 for groundedness based on whether the response uses the fact.
\end{enumerate}}

\textit{Example: }

\textit{Conversation History: \{\{Conversation\}\}}

\textit{Corresponding Fact: \{\{Contextual Fact\}\}}

\textit{Response: \{\{Generated Response\}\}}

\textit{First Assistant's Evaluation: \{\{Peer\_response1\}\}}

\textit{Second Assistant's Evaluation: \{\{Peer\_response2\}\}}

\textit{Third Assistant's Evaluation: \{\{Peer\_response3\}\}}

\textit{Evaluation Form (Answer by starting with ``Analysis:'' to analyze the given example regarding the evaluation criteria as concisely as possible, and then give the numeric rating on the next line by ``Rating''.)}

\textit{Groundedness:}
\end{mdframed}

\begin{mdframed}[style=MyFrame, frametitle={Naturalness Peer Prompt}]

\textit{You will be given a conversation between two individuals. You will then be given one potential response for the next turn in the conversation. The response concerns an interesting fact, which will be provided as well.}

\textit{Your task is to rate the responses on one metric.}
\textit{
Please make sure you read and understand these instructions carefully. Please keep this document open while reviewing, and refer to it as needed.}

\textit{Evaluation Criteria:}

\textit{Naturalness (1-3) Is the response naturally written??\\
- A score of 1 (bad) means that the response is unnatural.\\
- A score of 2 (ok) means the response is strange, but not entirely unnatural.\\
- A score of 3 (good) means that the response is natural.\\
}

\textit{Evaluation Steps:}
\textit{\begin{enumerate}
    \item Read the conversation between the two individuals.
    \item Read the potential response for the next turn in the conversation.
    \item Evaluate the response based on its naturalness, using the provided criteria.
    \item Assign a rating score of 1, 2, or 3 based on the evaluation.
\end{enumerate}}

\textit{Example: }

\textit{Conversation History: \{\{Conversation\}\}}

\textit{Corresponding Fact: \{\{Contextual Fact\}\}}

\textit{Response: \{\{Generated Response\}\}}

\textit{Evaluation Form (Answer by starting with ``Analysis:'' to analyze the given example regarding the evaluation criteria as concise as possible, and then give the numeric rating on the next line by ``Rating''.)}

\textit{Naturalness:}

\end{mdframed}

\begin{mdframed}[style=MyFrame, frametitle={Naturalness AreaChair Prompt}]
\textit{Navigate through a simulated conversation between two individuals, followed by a provided potential response incorporating an intriguing fact. Your role is to assess the responses based on the naturalness metric.}

\textit{Alongside your evaluation, you will also receive initial evaluations from three large language models, referred to as the assistants' evaluations. Please read the instructions and criteria below carefully and use them as a guide in your evaluation, critically assessing the conversation, and the assistants' inputs.}

\textit{Ensure a meticulous understanding of the instructions. Keep this document accessible for reference during the evaluation.}

\textit{Evaluation Criteria:}

\textit{Naturalness (1-3): Is the response naturally written?\\}
\textit{- A score of 1 (bad) means that the response is unnatural.\\
- A score of 2 (ok) means the response is strange, but not entirely unnatural.\\
- A score of 3 (good) means that the response is natural.\\
}
\\

\textit{Evaluation Steps:}
\textit{\begin{enumerate}
    \item Read the conversation between the two individuals.
    \item Read the potential response for the next turn in the conversation.
    \item Evaluate the response based on its naturalness, using the provided criteria.
    \item Assign a rating score of 1, 2, or 3 based on the evaluation.
\end{enumerate}}

\textit{Example: }

\textit{Conversation History: \{\{Conversation\}\}}

\textit{Corresponding Fact: \{\{Contextual Fact\}\}}

\textit{Response: \{\{Generated Response\}\}}

\textit{First Assistant's Evaluation: \{\{Peer\_response1\}\}}

\textit{Second Assistant's Evaluation: \{\{Peer\_response2\}\}}

\textit{Third Assistant's Evaluation: \{\{Peer\_response3\}\}}

\textit{Evaluation Form (Answer by starting with ``Analysis:'' to analyze the given example regarding the evaluation criteria as concisely as possible, and then give the numeric rating on the next line by ``Rating''.)}

\textit{Naturalness:}
\end{mdframed}

\subsection{MultiModal Evaluation}

\subsubsection{\textbf{ICQD}}

\begin{mdframed}[style=MyFrame, frametitle={Caption Quality Peer Prompt}]
\textit{Your task is to carefully evaluate the alignment between an image and its corresponding caption based on the provided criteria. Pay close attention to the instructions to ensure an accurate and nuanced assessment.}

\textit{Instructions:}
\textit{\begin{enumerate}
    \item Examine the image closely, identifying its key visual elements, objects, actions, and overall context.
    \item Scrutinize the caption, comparing it to the visual content of the image, and identifying any inaccuracies, omissions, or misleading information. Consider both the explicit details and the overall context of the image.
    \item Rate the caption on a scale of 1-100 according to the Evaluation Criteria, where 1 indicates a very poor match and 100 indicates a perfect match.
\end{enumerate}}

\textit{Evaluation Criteria:}

\textit{Rating (0-100): Evaluate the extent to which the caption aligns with the visual content of the image. A high rating should be given if the caption accurately reflects the main elements, actions, and context of the image, even if it uses concise language or omits minor details. Deduct points for inaccuracies, misleading descriptions, or significant omissions that distort the intended message of the image.}\\

\textit{- 90-100: The caption perfectly or almost perfectly captures the image's content.\\
- 70-89: The caption is mostly accurate, with only minor inaccuracies or omissions.\\
- 50-69: The caption has notable inaccuracies or omissions but still partially represents the image.\\
- 30-49: The caption poorly represents the image, with significant inaccuracies or misleading elements.\\
- 0-29: The caption is almost entirely inaccurate or irrelevant to the image.\\}

\textit{Example:}

\textit{Image:}

\textit{[Image will be provided separately]}

\textit{Caption: \{\{Caption\}\}}

\textit{Evaluation Form (Answer by starting with ``Analysis:'' to analyze the provided example regarding the evaluation criteria as concisely as possible, and then give the numeric rating on the next line by ``Rating''.)}

\textit{Caption\_Quality:}
\end{mdframed}

\begin{mdframed}[style=MyFrame, frametitle={Caption Quality AreaChair Prompt}]
\textit{You will be given an image, its caption, and you will also receive initial evaluations from two large language models, referred to as the assistants' evaluations.}

\textit{Your task is to rate the caption on one metric.}

\textit{Please read the instructions and criteria below carefully and use them as a guide in your evaluation.}

\textit{Evaluation Criteria:}

\textit{Relevance (0-100) - Assess how well the caption aligns with the content of the image. The caption should accurately describe or complement the visual elements and context of the image. Consider if the caption captures the key aspects of the image, its mood, and its intent, and whether it adds value by enhancing the viewer's understanding or experience of the image.}

\textit{Evaluation Guidelines:}
\begin{enumerate}[left=0pt, noitemsep]
    \item \textit{Examine the Image: Carefully observe the image to understand its main elements, context, and message.}
    \item \textit{Review the Caption: Analyze if the caption accurately and effectively describes or complements the image. Consider the appropriateness of the language, tone, and whether the caption adds meaningful context or insight.}
    \item \textit{Rate the Caption's Relevance on a Scale of 0 to 100:}
    \begin{itemize}[left=0pt, noitemsep]
        \item[-] \textit{90-100: The caption is highly relevant, fully capturing the essence of the image with precise and insightful description or commentary, adding significant value to the image.}
        \item[-] \textit{80-89: The caption is mostly relevant, capturing most key elements of the image with minor omissions or slightly less impactful language, still adding clear value.}
        \item[-] \textit{70-79: The caption is somewhat relevant, capturing some key aspects but missing others, or includes minor irrelevant details, with a noticeable but limited enhancement to the image.}
        \item[-] \textit{50-69: The caption has limited relevance, covering only a few elements of the image or providing a description that is either too generic or somewhat off-target, adding minimal value.}
        \item[-] \textit{30-49: The caption is marginally relevant, with significant omissions or inaccuracies, possibly detracting from the image by misrepresenting it or providing little to no useful context.}
        \item[-] \textit{10-29: The caption is largely irrelevant, missing the key aspects of the image, with significant inaccuracies or misrepresentations, adding no value or even confusing the viewer.}
        \item[-] \textit{0-9: The caption is completely irrelevant or nonsensical, with no connection to the image, possibly confusing or misleading the viewer.}
    \end{itemize}
\end{enumerate}

\textit{Example:}

\textit{Image:}

\textit{[Image is attached below]}

\textit{Caption: \{\{Caption\}\}}

\textit{First Assistant's Evaluation: \{\{Peer\_Response1\}\}\\
Second Assistant's Evaluation: \{\{Peer\_Response2\}\}}

\textit{Evaluation Form (Answer by starting with ``Analysis:'' to analyze the provided example regarding the evaluation criteria, incorporating the peer ratings, and then give the numeric rating on the next line by ``Rating''.)}

\textit{Caption\_Quality:}
\end{mdframed}

\subsubsection{\textbf{AGIQA}}

\begin{mdframed}[style=MyFrame, frametitle={Image Quality Peer Prompt}]
\textit{You will be given an image generated based on an input prompt.}

\textit{Your task is to rate the image on one metric.}

\textit{Please make sure you read and understand these instructions carefully. Please keep this document open while reviewing, and refer to it as needed.}

\textit{Evaluation Criteria:}

\textit{Image\_Quality (0-5) - the overall visual coherence and alignment with the input prompt. This rating should reflect how well the image matches the prompt, considering the clarity, relevance, and composition of the image.}

\textit{Evaluation Steps:}

\begin{itemize}[left=0pt, noitemsep]
    \item \textit{Review the "Input Prompt" carefully to understand the intended content, theme, and style.}
    \item \textit{Examine the generated image and compare it to the "Input Prompt". Check if the image accurately represents the prompt, is visually clear, and if the composition aligns with the expected outcome.}
    \item \textit{Assign a score for Image Quality on a scale of 0 to 5, where 0 is the lowest and 5 is the highest based on the Evaluation Criteria.}
\end{itemize}

\textit{Example:}

\textit{Input Prompt: \{\{Input\_Prompt\}\}}

\textit{Generated Image:}

\textit{[Image is attached below]}

\textit{Evaluation Form (Answer by starting with ``Analysis:'' to analyze the given example regarding the evaluation criteria as concise as possible, and then give the numeric rating on the next line by ``Rating''.)}

\textit{- Image\_Quality:}
\end{mdframed}

\begin{mdframed}[style=MyFrame, frametitle={Image Quality AreaChair Prompt}]
\textit{You will be given an image generated based on an input prompt, along with initial evaluations from two assistants, referred to as the assistants' evaluations.}

\textit{Your task is to rate the image on one metric.}

\textit{Please read the instructions and criteria below carefully and use them as a guide in your evaluation.}

\textit{Evaluation Criteria:}

\textit{Image\_Quality (0-5) - Assess the visual coherence and alignment of the image with the input prompt. The image should reflect the content, theme, and style described in the prompt, and be visually clear and well-composed.}

\textit{Evaluation Guidelines:}

\begin{itemize}[left=0pt, noitemsep]
    \item \textit{Review the "Input Prompt" to understand the intended content, theme, and style.}
    \item \textit{Examine the generated image and analyze how well it represents the "Input Prompt" in terms of accuracy, clarity, and composition.}
    \item \textit{Rate the image's quality on a scale of 0 to 5, with 0 being the lowest quality and 5 being the highest quality.}
    \item \textit{Scoring Guidelines:}
    \begin{itemize}[left=0pt, noitemsep]
        \item \textit{Score 5.0: The image fully captures the essence of the prompt with a high level of accuracy, clarity, and visual appeal, without any significant errors or irrelevant elements.}
        \item \textit{4 $\leq$ Score $<$ 5: The image mostly aligns with the prompt, with minor inaccuracies or less relevant details, but still maintains a generally high quality.}
        \item \textit{3 $\leq$ Score $<$ 4: The image partially represents the prompt, with noticeable inaccuracies or irrelevant details, and a less coherent visual presentation.}
        \item \textit{2 $\leq$ Score $<$ 3: The image has significant deviations from the prompt, with major inaccuracies, irrelevant elements, and a disjointed visual composition.}
        \item \textit{1 $\leq$ Score $<$ 2: The image fails to represent the prompt accurately, lacks visual coherence, and includes significant errors or irrelevant elements.}
        \item \textit{0 $\leq$ Score $<$ 1: The image is completely unrelated to the prompt.}
    \end{itemize}
\end{itemize}

\textit{Example:}

\textit{Input Prompt: \{\{Input\_Prompt\}\}}

\textit{Generated Image:}

\textit{[Image is attached below]}

\textit{First Assistant's Evaluation: \{\{Peer\_response1\}\}}

\textit{Second Assistant's Evaluation: \{\{Peer\_response2\}\}}

\textit{Please provide your analysis and rating as follows:}

\textit{Evaluation Form (Answer by starting with ``Analysis:'' to analyze the given example regarding the evaluation criteria as concise as possible, and then give the numeric rating on the next line by ``Rating''.)}

\textit{- Image\_Quality:}

\end{mdframed}

\subsection{Reasoning}

\subsubsection{\textbf{AQuA}}

\begin{mdframed}[style=MyFrame, frametitle={AQuA Peer Prompt}]
\textit{You will be provided with a problem that requires logical reasoning, mathematical calculation, or both.}

\textit{Your task is to solve the problem accurately, providing not just the correct answer but also a clear explanation of the steps taken to reach that answer.}

\textit{It is crucial to thoroughly understand the problem and apply the correct principles or formulas to solve it.}

\textit{Instructions:}

\begin{itemize}[left=0pt, noitemsep]
    \item \textit{Read the problem statement carefully, ensuring you understand all the details and what is required for the solution.}
    \item \textit{Work through the problem logically and methodically, explaining your reasoning and the steps you take to solve the problem.}
    \item \textit{Provide the final answer clearly, specifying it by choosing one of the provided options (e.g., A, B, C, etc.).}
\end{itemize}

\textit{Problem Statement:} \textit{\{\{question\}\}}

\textit{Provided Options:} \textit{\{\{options\}\}}

\textit{Evaluation Form:}
\begin{itemize}[left=0pt, noitemsep]
    \item \textit{Analysis: [Start with ``Analysis:'' to provide a concise and structured explanation of the steps and reasoning used to solve the problem. Ensure your analysis is clear and follows a logical sequence.]}
    \item \textit{Answer: [Clearly state the final answer only (e.g., A, B, C, etc.) on the line after your analysis.]}
\end{itemize}

\end{mdframed}

\begin{mdframed}[style=MyFrame, frametitle={AQuA area chair Prompt}]
\textit{You will be provided with a problem that requires logical reasoning, mathematical calculation, or both. Along with the problem, you will also receive solutions from three other Language Models (LLMs).}

\textit{Your task is to solve the problem accurately, using the peer responses to inform your approach. Apply the correct principles or formulas to arrive at the solution, while taking note of any useful insights or errors in the peer responses.}

\textit{Instructions:}

\begin{itemize}[left=0pt, noitemsep]
    \item \textit{Understand the Problem: Read the problem statement carefully, ensuring you grasp all details.}
    \item \textit{Review Peer Responses: Consider the solutions provided by the LLMs, noting useful approaches or any errors.}
    \item \textit{Solve the Problem: Work through the problem logically, explaining your reasoning and steps. Utilize the peer responses as needed but ensure your solution is accurate and complete.}
    \item \textit{Final Answer: Clearly state the final answer, choosing one of the provided options (e.g., A, B, C, etc.).}
\end{itemize}

\textit{Problem Statement:} \textit{\{\{question\}\}}

\textit{Provided Options:} \textit{\{\{options\}\}}

\textit{Solutions by Other LLMs:}

\begin{itemize}[left=0pt, noitemsep]
    \item \textit{LLM 1 Answer: \{\{Peer\_response1\}\}}
    \item \textit{LLM 2 Answer: \{\{Peer\_response2\}\}}
    \item \textit{LLM 3 Answer: \{\{Peer\_response3\}\}}
\end{itemize}

\textit{Evaluation Form:}

\begin{itemize}[left=0pt, noitemsep]
    \item \textit{Analysis: [Start with ``Analysis:'', provide a concise explanation of your reasoning and steps, integrating relevant insights from the LLMs’ responses.]}
    \item \textit{Answer: [Clearly state the final answer label ONLY (e.g., A, B, C, etc.) on the line after your analysis. (DO NOT GIVE ANYTHING ELSE).]}
\end{itemize}
\end{mdframed}

\subsubsection{\textbf{BBH\_DU}}

\begin{mdframed}[style=MyFrame, frametitle={BBH\_DU Peer Prompt}]

\textit{You will be provided with a problem that requires understanding and interpreting dates or times logically.}

\textit{Your task is to solve the problem accurately, providing not just the correct answer but also a clear explanation of the steps taken to reach that answer.}

\textit{It is crucial to thoroughly understand the problem, applying the correct principles or formulas to arrive at the solution.}

\textit{Instructions:}

\begin{itemize}[left=0pt, noitemsep]
    \item \textit{Read the problem statement carefully, ensuring you understand all the details and what is required for the solution.}
    \item \textit{Work through the problem logically and methodically, explaining your reasoning and the steps you take to solve the problem.}
    \item \textit{Provide the final answer clearly, specifying it by choosing one of the provided options (e.g., A, B, C, etc.).}
\end{itemize}

\textit{Problem Statement:} \textit{\{\{question\}\}}

\textit{Evaluation Form:}

\begin{itemize}[left=0pt, noitemsep]
    \item \textit{Analysis: [Start with ``Analysis:'' to provide a concise and structured explanation of the steps and reasoning used to solve the problem. Ensure your analysis is clear and follows a logical sequence.]}
    \item \textit{Answer: [Clearly state the final answer only (e.g., A, B, C, etc.) on the line after your analysis.]}
\end{itemize}
\end{mdframed}

\begin{mdframed}[style=MyFrame, frametitle={BBH\_DU AreaChair Prompt}]

\textit{You will be provided with a problem that requires understanding and interpreting dates or times logically. You will also receive the final answers from three other Language Models (LLMs).}

\textit{Your task is to solve the problem accurately, using the answers provided by the LLMs to inform your reasoning. Provide a clear explanation of your approach, and arrive at your own final answer.}

\textit{Instructions:}

\begin{itemize}[left=0pt, noitemsep]
    \item \textit{Understand the Problem: Read the problem statement carefully, ensuring you grasp all details.}
    \item \textit{Review Peer answers: Consider the final answers provided by the LLMs, noting any patterns or outliers.}
    \item \textit{Solve the Problem: Work through the problem logically, explaining your reasoning and steps. Use the peer answers as a reference but ensure your solution is accurate and complete.}
    \item \textit{Final Answer: Clearly state the final answer, choosing one of the provided options (e.g., A, B, C, etc.).}
\end{itemize}

\textit{Problem Statement:} \textit{\{\{question\}\}}

\textit{Answers from Other LLMs:}

\begin{itemize}[left=0pt, noitemsep]
    \item \textit{LLM 1 answer: \{\{Peer\_response1\}\}}
    \item \textit{LLM 2 answer: \{\{Peer\_response2\}\}}
    \item \textit{LLM 3 answer: \{\{Peer\_response3\}\}}
\end{itemize}

\textit{Evaluation Form:}

\begin{itemize}[left=0pt, noitemsep]
    \item \textit{Analysis: [Start with ``Analysis:'' to provide a concise explanation of your reasoning and steps to solve the problem, using the peer answers as a reference.]}
    \item \textit{Answer: [Clearly state the final answer label ONLY (e.g., A, B, C, etc.) on the line after your analysis. (DO NOT GIVE ANYTHING ELSE).]}
\end{itemize}
\end{mdframed}

\subsubsection{\textbf{CSQA}}
\begin{mdframed}[style=MyFrame, frametitle={CSQA Peer Prompt}]

\textit{Evaluate the question by selecting the best option from the provided choices. Your task is to understand the context and nuances of the question, utilize your knowledge of the topic, and determine the most appropriate answer based on the options given. The goal is to select the most relevant and correct option that aligns with the question's intent.}

\textit{Instructions:}

\begin{itemize}[left=0pt, noitemsep]
    \item \textit{Understand the Question: Read the question carefully to comprehend all aspects and the context in which it is asked.}
    \item \textit{Consider the Options: Analyze each provided option carefully. Think about how each option relates to the question and the scenario it presents.}
    \item \textit{Select the Best Option: Choose the option that best answers the question, based on your analysis. Focus on the logic or knowledge that supports this choice.}
\end{itemize}

\textit{Problem Statement:} \textit{\{\{question\}\}}

\textit{Provided Options:} \textit{\{\{options\}\}}

\textit{Evaluation Form:}

\begin{itemize}[left=0pt, noitemsep]
    \item \textit{Analysis: [Begin with ``Analysis:'' to provide a structured and clear explanation of your reasoning process. Your analysis should logically explain why the chosen option is the most appropriate answer to the question.]}
    \item \textit{Answer: [Clearly state the final answer only (e.g., A, B, C, etc.) on the line after your analysis.]}
\end{itemize}
\end{mdframed}

\begin{mdframed}[style=MyFrame, frametitle={CSQA area chair Prompt}]

\textit{You will be provided with a question that requires careful evaluation to select the best option from the provided choices. You will also receive the final answers from three other Language Models (LLMs).}

\textit{Your task is to determine the most appropriate answer, using the answers provided by the LLMs to inform your reasoning. Provide a clear explanation of your thought process and select the option that best aligns with the question's intent.}

\textit{Instructions:}

\begin{itemize}[left=0pt, noitemsep]
    \item \textit{Understand the Question: Read the question carefully to comprehend all aspects and context.}
    \item \textit{Review Peer Answers: Consider the final answers provided by the LLMs, noting any patterns or outliers.}
    \item \textit{Select the Best Option: Based on your understanding and the peer answers, choose the option that best answers the question.}
    \item \textit{Final Answer: Clearly state the final answer, choosing one of the provided options (e.g., A, B, C, etc.).}
\end{itemize}

\textit{Problem Statement:} \textit{\{\{question\}\}}

\textit{Provided Options:} \textit{\{\{options\}\}}

\textit{Answers from Other LLMs:}

\begin{itemize}[left=0pt, noitemsep]
    \item \textit{LLM 1 Answer: \{\{Peer\_response1\}\}}
    \item \textit{LLM 2 Answer: \{\{Peer\_response2\}\}}
    \item \textit{LLM 3 Answer: \{\{Peer\_response3\}\}}
\end{itemize}

\textit{Evaluation Form:}

\begin{itemize}[left=0pt, noitemsep]
    \item \textit{Analysis: [Start with ``Analysis:'' to provide a concise and clear explanation of your reasoning, using the peer answers as a reference.]}
    \item \textit{Answer: [Clearly state the final answer label ONLY (e.g., A, B, C, etc.) on the line after your analysis. (DO NOT GIVE ANYTHING ELSE).]}
\end{itemize}
\end{mdframed}

\subsubsection{\textbf{GSM8k}}
\begin{mdframed}[style=MyFrame, frametitle={GSM8k Peer Prompt}]

\textit{You will be provided with a problem that requires logical reasoning, mathematical calculation, or both.}

\textit{Your task is to solve the problem accurately, providing not just the answer but also a clear explanation of the steps taken to reach that answer.}

\textit{It is crucial to understand the problem thoroughly and apply the correct principles or formulas to solve it.}

\textit{Instructions:}

\begin{itemize}[left=0pt, noitemsep]
    \item \textit{Read the problem statement carefully, ensuring you understand all the details and what is required for the solution.}
    \item \textit{Work through the problem logically and methodically, explaining your reasoning and the steps you take to solve the problem.}
    \item \textit{Provide the final answer clearly, specifying it as a numerical value or a specific explanation as required by the problem statement.}
\end{itemize}

\textit{Problem Statement:} \textit{\{\{question\}\}}

\textit{Evaluation Form:}

\begin{itemize}[left=0pt, noitemsep]
    \item \textit{Analysis: [Start with ``Analysis:'' to provide a concise and structured explanation of the steps and reasoning used to solve the problem. Ensure your analysis is clear and follows a logical sequence.]}
    \item \textit{Answer: [Clearly state the final answer only (number) on the line after your analysis.]}
\end{itemize}
\end{mdframed}

\begin{mdframed}[style=MyFrame, frametitle={GSM8k area chair Prompt}]

\textit{You will be provided with a problem that requires logical reasoning, mathematical calculation, or both. You will also receive the final answers from three other Language Models (LLMs).}

\textit{Your task is to solve the problem accurately, using the peer answers to inform your reasoning. Provide a clear explanation of your thought process and the steps taken to arrive at the solution. Ensure that your reasoning is sound and the final answer is correct.}

\textit{Instructions:}

\begin{itemize}[left=0pt, noitemsep]
    \item \textit{Understand the Problem: Read the problem statement carefully to ensure you grasp all the details and what is required.}
    \item \textit{Review Peer Answers: Consider the final answers provided by the LLMs, noting any patterns or outliers.}
    \item \textit{Work Through the Problem: Solve the problem methodically, using the peer answers as a reference. Explain your reasoning clearly.}
    \item \textit{Final Answer: Provide the final answer clearly, specifying it as a numerical value or as required by the problem statement.}
\end{itemize}

\textit{Problem Statement:} \textit{\{\{question\}\}}

\textit{Answers from Other LLMs:}

\begin{itemize}[left=0pt, noitemsep]
    \item \textit{LLM 1 Answer: \{\{Peer\_response1\}\}}
    \item \textit{LLM 2 Answer: \{\{Peer\_response2\}\}}
    \item \textit{LLM 3 Answer: \{\{Peer\_response3\}\}}
\end{itemize}

\textit{Evaluation Form:}

\begin{itemize}[left=0pt, noitemsep]
    \item \textit{Analysis: [Start with ``Analysis:'' to provide a concise and clear explanation of your reasoning, using the peer answers as a reference.]}
    \item \textit{Answer: [Clearly state the final answer ONLY (number) on the line after your analysis. (DO NOT GIVE ANYTHING ELSE).]}
\end{itemize}
\end{mdframed}

\end{document}